\documentclass[]{interact}
\usepackage[table]{xcolor}
\usepackage{hyperref}
\usepackage{graphicx}
\usepackage{float}
\usepackage{subcaption}
\usepackage{xcolor}
\usepackage{epstopdf}

\usepackage{natbib}
\usepackage{booktabs}
\usepackage{pifont}
\usepackage{caption}
\usepackage{array}
\usepackage{verbatim}
\usepackage{amsmath}

\usepackage{comment}

\bibpunct[, ]{(}{)}{;}{a}{}{,}
\renewcommand\bibfont{\fontsize{10}
{12}\selectfont}
\newcommand{\xmark}{\ding{55}} 
\newcommand{\cmark}{\ding{51}} 

\newcolumntype{L}[1]{>{\raggedright\arraybackslash}p{#1}} 
\newcolumntype{C}[1]{>{\centering\arraybackslash}p{#1}} 

\theoremstyle{plain}

\theoremstyle{definition}

\theoremstyle{remark}

\begin{document}

\articletype{Preprint: Under Review}


\title{Agentic LLMs in the Supply Chain: Towards Autonomous Multi-Agent Consensus-Seeking}

\author{
\name{Valeria Jannelli\textsuperscript{a,b}\thanks{CONTACT A. Brintrup. Email: ab702@cam.ac.uk}, Stefan Schöpf\textsuperscript{b,c}, Matthias Bickel\textsuperscript{a}, Torbjørn Netland\textsuperscript{a}, \\  and Alexandra Brintrup\textsuperscript{b,c}}
\affil{\textsuperscript{a}ETH Zurich, Zurich, Switzerland \\
\textsuperscript{b}University of Cambridge, Cambridge, United Kingdom \\
\textsuperscript{c}The Alan Turing Institute, London, United Kingdom}
}

\maketitle
\begin{abstract}
This paper explores how Large Language Models (LLMs) can automate consensus-seeking in supply chain management (SCM), where frequent decisions on problems such as inventory levels and delivery times require coordination among companies. 
Traditional SCM relies on human consensus in decision-making to avoid emergent problems like the bullwhip effect. Some routine consensus processes, especially those that are time-intensive and costly, can be automated.
 Existing solutions for automated coordination have faced challenges due to high entry barriers locking out SMEs, limited capabilities, and limited adaptability in complex scenarios. However, recent advances in Generative AI, particularly LLMs, show promise in overcoming these barriers. LLMs, trained on vast datasets can negotiate, reason, and plan, facilitating near-human-level consensus at scale with minimal entry barriers. 
In this work, we identify key limitations in existing approaches and propose autonomous LLM agents to address these gaps. We introduce a series of novel, supply chain-specific consensus-seeking frameworks tailored for LLM agents and validate the effectiveness of our approach through a case study in inventory management. To accelerate progress within the SCM community, we open-source our code, providing a foundation for further advancements in LLM-powered autonomous supply chain solutions. \\ 
SDG 9: Industry, innovation and infrastructure

\end{abstract}


\newpage

\section{Introduction}

Supply chain management (SCM) requires continuous decision-making during day-to-day operations, on a multitude of problems ranging from inventory planning to delivery scheduling \citep{christopher2016logistics}.
As supply chains are interdependent systems of multiple companies, these decisions often need to be made via consensus between self-interested companies. Here, consensus-seeking is defined as a process that involves multiple parties interacting to reach a decision, whereby each party, in our case a supply chain firm, has different beliefs and goals. However, all parties have a common interest in reaching an agreement on selecting the best decision. Within the context of supply chains, consensus-seeking is closely related to supply chain coordination - defined as “collaborative working for joint planning, joint product development, mutual exchange of information and integrated information systems, cross coordination on several levels in the companies on the network, long-term cooperation and fair sharing of risks and benefits" \citep{skjoett2000european}. A common example of consensus-seeking occurs in demand planning, where participants along a chain need to agree on appropriate order quantities. Failure to achieve consensus results in the bullwhip effect, where small demand changes can cause large production fluctuations  \citep{lee1997bullwhip, lee1997information}. Coordination and consensus form a rich part of supply chain management research, from a range of  disciplinary input and discourse on information asymmetry and information sharing, planning and uncertainty, the role of technology and other coordination instruments such as contracts.  

\begin{figure}
    \centering
    \includegraphics[width=0.40\textwidth]{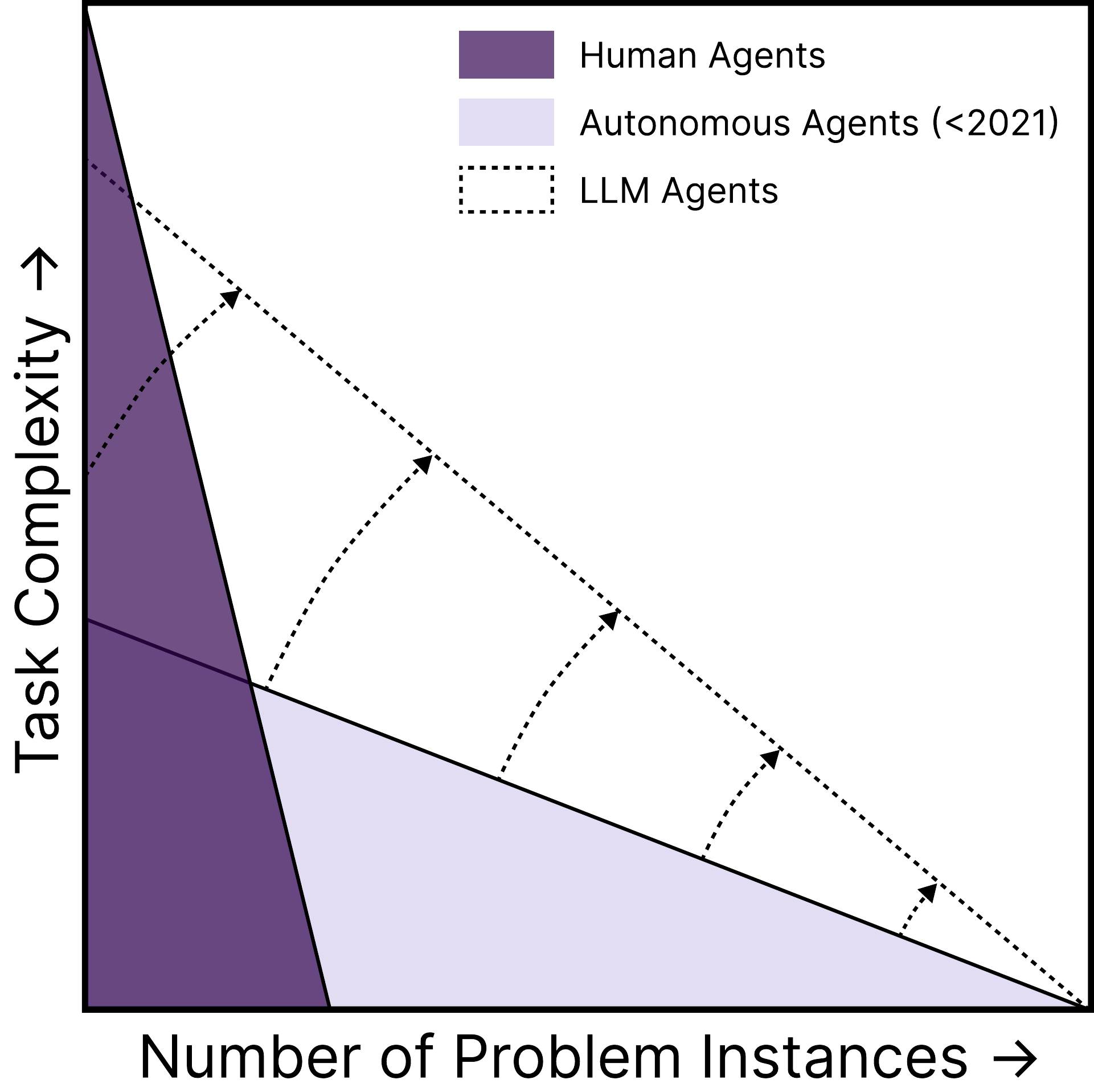}
    \caption{Complex consensus-seeking tasks are out of scope for current autonomous agents. LLM agents promise a new frontier, enabling complex consensus finding at scale beyond human speeds and thus unlocking significant efficiency improvements.}
    \label{fig:consensus}
\end{figure}

This paper focuses on a subset of supply chain consensus-seeking problems, that could benefit from automated handling using AI. Decisions on delivery quantities, order frequency, and capacity allocation are given frequently and often involve situations where end-to-end coordination would yield superior solution outcomes \citep{towards-autonomous-sc, viswanadham2002past}. However end-to-end coordination requires manual orchestration, which in turn requires time and resources to be allocated to consensus building among self-interested supply chain actors. While humans are able to solve complex consensus tasks, they are limited in the volume of tasks they can fulfil in a given time window as illustrated in Fig. \ref{fig:consensus}. End-to-end consensus itself is a particularly difficult task that involves multiple actors to iteratively communicate with one another to achieve consensus. Given the high complexity of doing so at scale, individual firms are disincentivised to take part in consensus-seeking, unless the relationship or the end goal “is worth it”. 

A lack of end-to-end consensus often results in small problems to aggregate and lead to sub-optimal outcomes, such as loss of efficiency (e.g., low utilization of capacity \citep{mak2023fair}) and information distortion (e.g., bullwhip effect \citep{lee1997bullwhip}, shortage gaming \citep{samuel2003shortage}). Some examples of issues that arise between companies include lack of timely information and inefficient crisis response due to ineffective communication exchange between supply chain players \citep{crisis-response-info-networks}. ~\citet{ma2019modelling} argues that more effective consensus-seeking and coordination should be aspired to in a world with increasing supply chain disruptions. 

Researchers have, in the past, proposed autonomous algorithms to handle low-level operational coordination in supply chains \citep{towards-autonomous-sc}. The idea behind these proposals was that automation would alleviate manual effort, therefore allowing end-to-end, systemic solution optimality. Decisions would have become more traceable and transparent.  Most of these proposals centred around intelligent software agents, which are computer programs that mimic humans and act on their behalf \citep{wooldridge199595}. They do so by perceiving their environment (e.g., ERP system data, digital twins), interacting with other agents (e.g., negotiation), using tools (i.e., other algorithms), and making decisions to achieve their given goals.
However as reflected by practice, these concepts did not gain traction in real-life SCM \citep{hendler2007all,marik2005industrial}. Reasons for failure are numerous but two key causes have been hypothesized:  
First, developing agent solutions requires expertise and curated data. Hence barriers to entry are too high due to costs and skill shortages \citep{towards-autonomous-sc}. Agents need significant adaptation to work on new tasks or changes in the environment (e.g., data drift, new objectives). Furthermore, agents need well-defined interaction protocols and communication templates to work with other automated entities or humans across companies, turning the problem into a multi-company change management project \citep{marik2005industrial}. Effectively, agents are not feasible to implement and run for SMEs with their available resource budgets - but SMEs are key to enable end-to-end coverage of supply chains to automate the vast majority of consensus-seeking tasks.

Second, agents struggle to achieve performance that is comparable to humans as consensus-seeking tasks become more complex and requires creative problem-solving. Achieving or surpassing human performance is possible in well-defined problem settings (e.g., games such as chess \citep{campbell2002deep} or Go \citep{chen2016evolution}) but intelligent agents fall short of human performance in complex settings that require reasoning or planning or are constantly changing and adapting \citep{bhakthavatsalam2021think_arc, grace2018will}.

The field of Generative AI (GenAI), which can generate content such as natural text or video after being trained on vast amounts of data, has seen enormous progress in recent years, especially in the GenAI branch of Large Language Models (LLMs) which can be used to generate text in numerous applications such as for chat interactions between autonomous agents \citep{jackson2024generative} or computer code for supply chain simulation \citep{jackson2024natural}.
Advancements in LLMs have shown logic and reasoning at near-human levels to address these two key challenges, possibly enabling a new frontier as shown in Fig. \ref{fig:consensus}.
First, LLMs already possess skills from pertaining on enormous datasets, can be easily accessed via APIs, and users can instruct them via an intuitive natural language interface. This allows companies to leverage LLMs as agents with minimal implementation and hosting effort. Existing databases can also be connected to enable LLMs to query them for additional information.
Second, recent works have also shown that LLMs as mediators \citep{mas-survey} or as negotiators \citep{welfare-diplomacy} can significantly improve consensus-seeking and overall outcome compared to human-human interaction, even when tackling highly contested topics. The flexibility of LLMs to reach a consensus with each other as well as with humans is especially relevant for real-life usage, as not all companies will have LLM agents, necessitating that an LLM agent is able to also interact with humans at those companies.

While LLM agents seem promising for SCM consensus-seeking at scale, to the best of our knowledge, no literature addresses this challenge from a perspective that incorporates the challenges of SCM such as sequential dependency of companies and partial observability of the whole system. We address this literature gap with a SCM-specific LLM agent framework and an accompanying case study (building upon the simulation environment of \cite{liu-2022}) to demonstrate its effectiveness in a sequential supply chain inventory replenishment task.
 Our case study shows that LLM agents are a promising step forward for SCM but managers need to be aware of their current limitations for practice.

This paper offers three contributions to the nascent literature on LLM-powered supply chain agents:
First, a modular communication framework for LLM-powered agents in the sequential supply chain, including the possibility for tool usage and communication between neighbouring agents.
Second, an empirical case study involving experiments across supply chain metrics and framework sophistication levels: Our results highlight in which situations using tools within the framework can achieve significant performance enhancements, and in which cases it is best to prioritize sophisticated multi-agent communication over tool usage.
Third, an open source implementation of the communication framework for the SC research community to use and build upon, already integrated in a sequential supply chain simulation environment that is problem-agnostic, scalable and can be extended to analyse similar topics.

\section{Related Work}

The global supply chain ecosystem is becoming increasingly complex and intricate, characterized by different product lines, supplier relationships, and scheduling needs. On top of this, disruptions in the supply chain are becoming  more frequent \citep{ivanov2023toward, bode2011understanding} due to increasing symptoms of climate change,  regional epidemics or global pandemics, as well as recurring regional conflicts. Due to the increasing scale and complexity of SCs, companies are facing difficulties in optimizing supply chain processes, as traditional optimization methods lack the ability to capture the complexity ad vulnerabilities of real-world SCs \citep{smyth2024artificial}. 

This state of affairs calls for faster, more automated, and more scalable supply chain consensus-seeking and coordination, that allows supply chain actors to improve their coordination and respond to frequent changes \citep{sc-coll-hot}. 

In the following, we discuss how multi-agent systems can be used to improve automated consensus-seeking and coordination in supply chain management research, the challenges with previous  approaches, as well as the benefits of introducing Generative AI technology, and LLMs in particular. We will also discuss why companies struggle with coordination and the role of technological support to achieve it. We look into the availability of benchmarking environments both in supply chain management research and in computer science research which we draw inspiration from to build our frameworks. 

In computer science, agents are defined as  computational entities in an environment from which they perceive different parameters that are used to make decisions, take  goal-directed decisions and actuate them ~\citep{mas-survey}. A Multi-Agent System (MAS) is comprised of multiple interacting agents with distinct goals and behaviours, working together or competing in a shared environment to carry out tasks \cite{wooldridge2000gaia}. In cooperative settings, agents consider each other’s information and decisions, aiming for consensus and optimal coordination ~\citep{mas-survey}. Each entity operates with limited information and must make decisions autonomously, while depending on the performance and actions of other entities in upstream and downstream tiers. These characteristics make MAS particularly suited to study sequential supply chains with multiple tiers of dependency and partial observability. In fact, a large number of multi-agent simulation studies in supply chains exist. However, we note that a MAS, as opposed to a simulation, is a system that hosts computational agents that interact with and impact real life.  
Therefore, a topic of interest is the degree of autonomy of the agents should have in a SCM environment ~\citet{towards-autonomous-sc}, which is defined as the degree to which a subset of agents has ``the capability to determine, conduct and control the actions or behaviours independently, without external input''. 

SC automation with agents have started with eProcurement \citep{neef2001procurement}, and led to automated negotiation \citep{jiao2006agent} and learning in business-to-business environments \citep{coehoorn2004learning}. Beginning in the 2000s, SCM researchers turned their attention to supply chain coordination using software agent technology, focusing on the efficient division of agent roles, to achieve decision optimality, an effective information sharing on end-to-end supply chains \citep{xue2005agent}. Works attempted to tackle a diverse range of problems such as inventory planning and order scheduling \citep{julka2002agent}, collaborative production planning \citep{dangelmaier2005supply}, automated procurement \citep{brintrup2011will}, disruption management \citep{behdani2019agent} across various industrial sectors such as aerospace, logistics, recycling, construction and food, and agriculture (see \cite{towards-autonomous-sc} for a review). 

The main way of achieving consensus in the software agent research in SCM, has been through optimization functions, either through a global objective function (e.g., \citealt{fung2005multiagent}), often pursued through a mediator agent (e.g., \citealt{pan2009optimal}), a pareto optimal solution search between conflicting objectives (e.g., \citealt{wong2010multi}), or ensuring that individual objectives do not conflict, by fine-tuning them a priori (e.g., \citealt{lim2013using}). A recent survey of these approaches by \cite{towards-autonomous-sc} argues that the field has stagnated due to low industrial adoption, which is partly due to the difficulty in creating bespoke, tailored solutions, as well as interoperability and system integration issues, leading to a lack of uptake from multiple supply chain partners. Lack of trust in automation and lack of explainability and decision traceability were other reasons cited. 

More recently, another agent based automation paradigm, Reinforcement Learning (RL) has gained attention in SCM. RL is a comprehensive AI framework, wherein agents equipped with RL capability determine how to act upon an environment to maximize a reward function \citep{boute2022deep}. The main applications of RL in supply chain management are information handling, transportation, and inventory management, RL agents coordinate material flows across sites \citep{rolf2023review}. In particular, Multi-Agent Reinforcement Learning (MARL) is a paradigm that studies the behavior of multiple agents coexisting and interacting with each other in a shared environment and is relevant to numerous industry scenarios \citep{si2024efficient, yang-rl-benchmark}.
Despite its broad applicability, MARL implementations are hampered by complex agent interactions and non-stationary dynamics resulting from both environment and the interactive learning agents \citep{yang-rl-benchmark}. A significant challenge in the applicability of MARL to a SCM setting is the “competitive" nature of supply chains, whereby agents are neither fully cooperative, working towards a single reward function, not they are fully competitive, working towards their own objective functions. There is an over-emphasis in AI research on MAS that engage in fully competitive zero-sum games or cooperative problems where AI agents attempt to jointly improve welfare \citep{prob-coop-ai}. 

A further challenge of MARL has to do with partial observability of the supply chain setting: independently learning agents with a local view struggle to reach optimal decisions, as the individual cannot obtain global information from its local observations. Moreover, policies change continuously due to all agents learning simultaneously \citep{survey-marl}. Thus, RL implementations have been reported to present significant challenges for real-world supply chain applications, especially for SMEs, due to the ``cold start'' problem: RL agents initially lack prior data or knowledge, requiring extensive interactions with the environment to gather useful data, which makes the learning process slow and inefficient. This challenge is compounded by the technical complexities of data collection, as observed by \citet{cannas2024artificial}. For machine learning and deep learning systems to learn effectively, they require access to large volumes of high-quality data; these are advantages that SMEs typically lack. 

Generative Artificial Intelligence (GenAI) has been designed to create new content, or data, by learning patterns from existing data. Unlike traditional Artificial Intelligence where the output depends on the given inputs, GenAI can generate novel outputs such as synthetic data generation \citep{zhang2018generative} and LLMs like ChatGPT \citep{openai2024release}, Copilot \citep{coploi2023genai}, Gemini \citep{gemini2023genai}, LLaMA \citep{touvron2023llama} used for the generating of natural language text, images, audio, and video. In this paper, we mainly focus on LLMs. 

With the advent of LLMs, new levels of automation and decision support in SCM are expected \citep{jackson2024generative}. Researchers have already demonstrated promising applications such as server placement optimization at Microsoft using LLMs \citep{li-sc-opt} and supply chain simulation creation from natural text using LLMs \citep{jackson2024natural}.

A recent development within the field of LLMs are LLM-powered agents \citep{towards-autonomous-sc}. These LLM-powered agents are able to interact with other agents or humans, use additional software (e.g., a numerical solver), query documents and databases, search the web, and more as research progresses rapidly.

Unlike previous generations of agents that required rigorously defined inputs and outputs, LLM-powered agents are capable of learning from vast amounts of unstructured data, enabling them to adapt to complex environments in real-time \citep{li-sc-opt}. Their ability to process natural language inputs and contextualize information allows them to act more autonomously, which is a crucial capability, especially in large sequential supply chains \citep{towards-autonomous-sc}. \\

LLM-powered agents also offer a promising alternative to RL approaches. These agents can leverage pre-trained models, allowing them to adapt quickly to new environments with minimal data \citep{zero-shot}. LLMs are particularly suited for applications where contextual understanding and decision-making are required without the need for large datasets, making them an attractive option for SMEs looking to integrate AI into their supply chains. For example, tool usage, which means adding external functions and solvers to steer the outputs of LLM-powered agents, is a crucial emerging research field to overcome LLM shortcomings, such as the fact that LLMs are pre-trained and therefore not fine-tuned for specific use cases ~\citep{wu-autogen-convo}. By integrating tools for specialized tasks, memory and/or Retrieval Augmented Generation in LLM-powered supply chain workflows, research is equipped with new strategies to orchestrate and optimize complex business decisions \citep{bcg_article}.

Advancements in LLM-powered multi-agent architectures for supply chain management have not yet found their way into industrial applications \citep{will-bots-take-over}. Furthermore, many systems for multi-agent communication developed in computer science research or supply chain management research are formulated as use cases, which are very context-specific and difficult to adapt to tailored situations. If communication systems were more easy-to-configure, this itself may promote more cooperation between members of real-world supply chains \citep{will-bots-take-over}.  

\citet{sc-coll-hot} find that companies tend to interact prevalently with companies both immediately upstream and immediately downstream in a supply chain, and are focused on improving sourcing, procurement, and supplier management processes. This reflects the partial observability of supply chains. Our goal is to investigate how LLMs can be used as a viable technology to improve collaboration in an end-to-end supply chain setting, as they offer a fast, natural language-based interface that is suited for industrial control \citep{song-industrial-control}.\\

\subsection{Available Benchmarking Environments in Supply Chain Management Research}

There is a lack of multi-agent LLM-powered benchmarking environments where longstanding challenges such as inventory management can be investigated. While there is a large corpus of existing research on AI applications for SC automation involving distributed agent architectures \citep{will-bots-take-over, liu-2022, survey-marl}, the supply chain research community does not have an SC-specific LLM-powered benchmarking environment with communication frameworks that reflects partial observability and a scalable sequential supply chain. 

While there is no SC-specific LLM-powered benchmarking environment, the field of Computer Science  has made numerous advancements in developing environments that study behaviours of cooperative and competitive LLM-powered agents \citep{zhao-competeai, welfare-diplomacy}. By drawing inspiration from LLM-powered decision-making frameworks from computer science research, as well as multi-agent supply chain environments that don't make use of LLMs, we can design a set of LLM-powered SC-specific consensus-seeking frameworks that reflect partial observability and collaboration between neighbours.

Examples of these environments developed by computer science research include: ~\citet{zhao-competeai}, which simulates competitive agent dynamics in a marketplace setting. Though this environment illustrates an industry example with partial observability, it lacks the sequential supply chain structure with partial observability that would be useful to study inventory management challenges. ~\citet{welfare-diplomacy} focus on multi-agent collaboration in a general-sum setting, which would also be interesting for LLM-powered supply chain interactions. However, a similar supply chain environment would have to include sequential dependencies, partial observability, as well as real-time decision-making and consensus-seeking. There is very limited research on LLM-powered decision-making in a supply chain setting, but \citet{quan2024invagent} introduces LLMs to manage inventory systems by leveraging zero-shot learning capabilities to minimize costs and stockouts. Despite testing Chain-of-Thought reasoning for LLM-powered agents in various supply chain scenarios, this implementation relies on standalone agents that don't negotiate with their neighbours to make their decisions.

Hence, another key contribution of this paper is an LLM-powered SC-specific consensus-seeking framework that focuses on varying types of interactions between neighbouring agents.

\section{Problem Setting}

We place our multi-agent communication frameworks in an environment that simulates an end-to-end supply chain inventory management setting, based on the work by ~\citet{liu-2022}. This environment has desirable features for our purposes such as partial observability, multiple tiers of dependency, as well as a general sum setting.  Our choice to focus our implementation on the end-to-end supply chain is due to the fact that this setting allows us to explore and tackle longstanding challenges in SCM research, such as the bullwhip effect, while allowing for a simple implementation of our communication framework that can be extended to a supply chain network. 

We implement our communication framework on top of ~\citet{liu-2022} because of its intuitive inventory management setting and built-in bullwhip effect metrics, as well as its easy extendability to a supply chain network setting with multiple parallel layers of echelons. 

\begin{figure}[h]
    \centering
    \includegraphics[trim={0cm 6cm 0cm 6cm}, clip, width=1.0\textwidth]{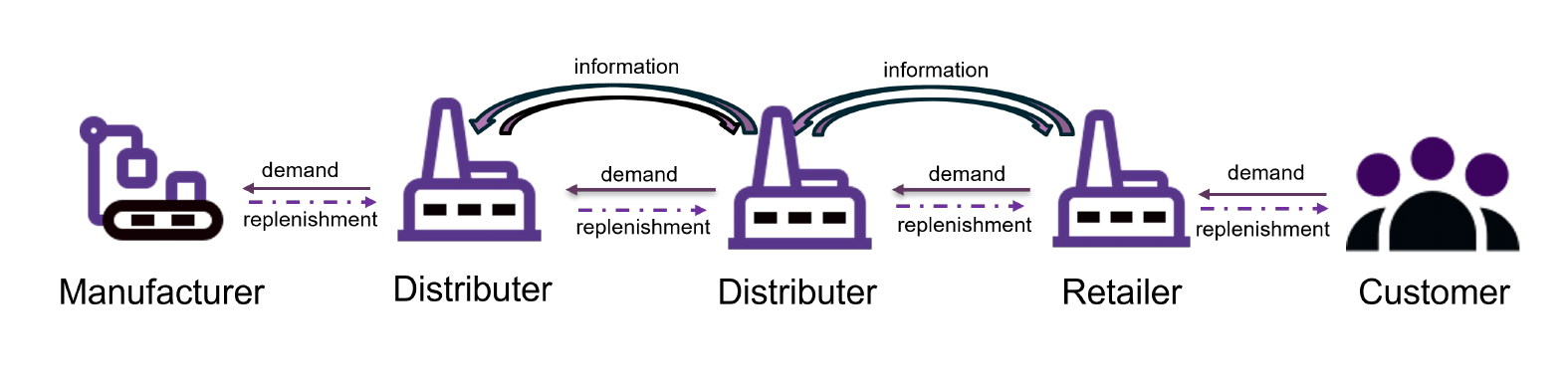}
    \caption{Problem setting: sequential supply chain with partial observability.}
    \label{problem-setting}
\end{figure}

Fig. \ref{problem-setting} shows a sequential supply chain that reflects the exchanges in order requests (``demand''), material products (``replenishment''), and additional insights (``information'') that flow between the communicating agents in our sequential supply chain setting. In the following, we describe the metrics that we will attempt to mitigate by using our increasingly sophisticated communication frameworks, as well as dedicated tools. 

\subsection{Global Costs in the end-to-end Supply Chain}
The global costs of our end-to-end supply chain are based on the implementation of local costs by ~\citet{liu-2022}. Specifically, the global costs are given by the sum of all the local costs of each agent, cumulated across each step of the inventory management simulation. The local costs of each agent are given by the sum of different cost components: inventory costs, backlog costs, where backlog is defined as the unmet demand from an agent's downstream neighbor, variable ordering cost, and fixed ordering cost.\\

\subsection{The Bullwhip Effect in end-to-end Supply Chains}
The bullwhip effect is a phenomenon of the increase in variability of orders as one moves upstream in a supply chain \citep{disney-bullwhip-2}.
It has been advocated that sharing information between neighbouring agents in the supply chain can mitigate the bullwhip effect ~\citep{bullwhip-progress-trends}. To compute the bullwhip effect in our sequential supply chain setting, we consider a formula derived by ~\citet{liu-2022} and based on \citet{fransoo-2000}, which considers the coefficient of variation of demand generated by a given echelon, i.e., the ratio between the standard deviation of demand and the mean of demand, as expressed by the following formula:\\\\
\[\text{coeff}_{var}=\frac{\sigma(demand_{\text{i}}(t, t+1))}{\mu(demand_{\text{i}}(t, t+1))}\]
\\\\
where demand is given by the list of historical actions of agent $i$, and $t$ is the current time interval. Therefore, a coefficient of variability of demand below 1 indicates that the bullwhip effect is negligible.  

To compute the aggregate bullwhip effect, we use the insight from ~\citet{fransoo-2000}, who conclude that the aggregate bullwhip effect for a (strictly) end-to-end supply chain can be achieved by multiplying all of the coefficients of variation (i.e., the bullwhip effects) of each individual echelon. 

In order to mitigate the bullwhip effect, we implement a traditional extant approach to order an optimal quantity based on SCM restocking policies. The metric used to optimize for the bullwhip effect is based on the Economic Order Quantity (EOQ) measure for each agent \citep{optim-methods-bw}, motivated by a similar approach which was adopted by \citet{jackson2024natural}. However, the challenge that arises is that the EOQ quantity picked by each individual agent may not be optimal for the entire supply chain. \citet{optim-methods-bw} and \citet{bullwhip-info-enrichment} underline that agreements on order amounts between neighbouring agents help mitigate the bullwhip effect. According to this insight, our implementation of communication frameworks based  negotiation should aid in mitigating the bullwhip effect. We deliberately adapt the original EOQ formula as a benchmark, and make it suitable for our environment \citep{liu-2022}, as follows:
\\
\[EOQ=\sqrt{\frac{2*\text{Demand}*\text{Ordering Cost}}{\text{Holding Cost}}}\]
\\
, where Demand is derived from the mean of historical demand data from the downstream agent, Ordering Cost is the per-unit cost of ordering, and Holding Cost is the per-unit cost of holding inventory.

\section{Methodology}

\begin{figure}[H]
    \centering
    \begin{subfigure}[t]{0.3\textwidth}
        \centering
        \includegraphics[width=0.85\textwidth]{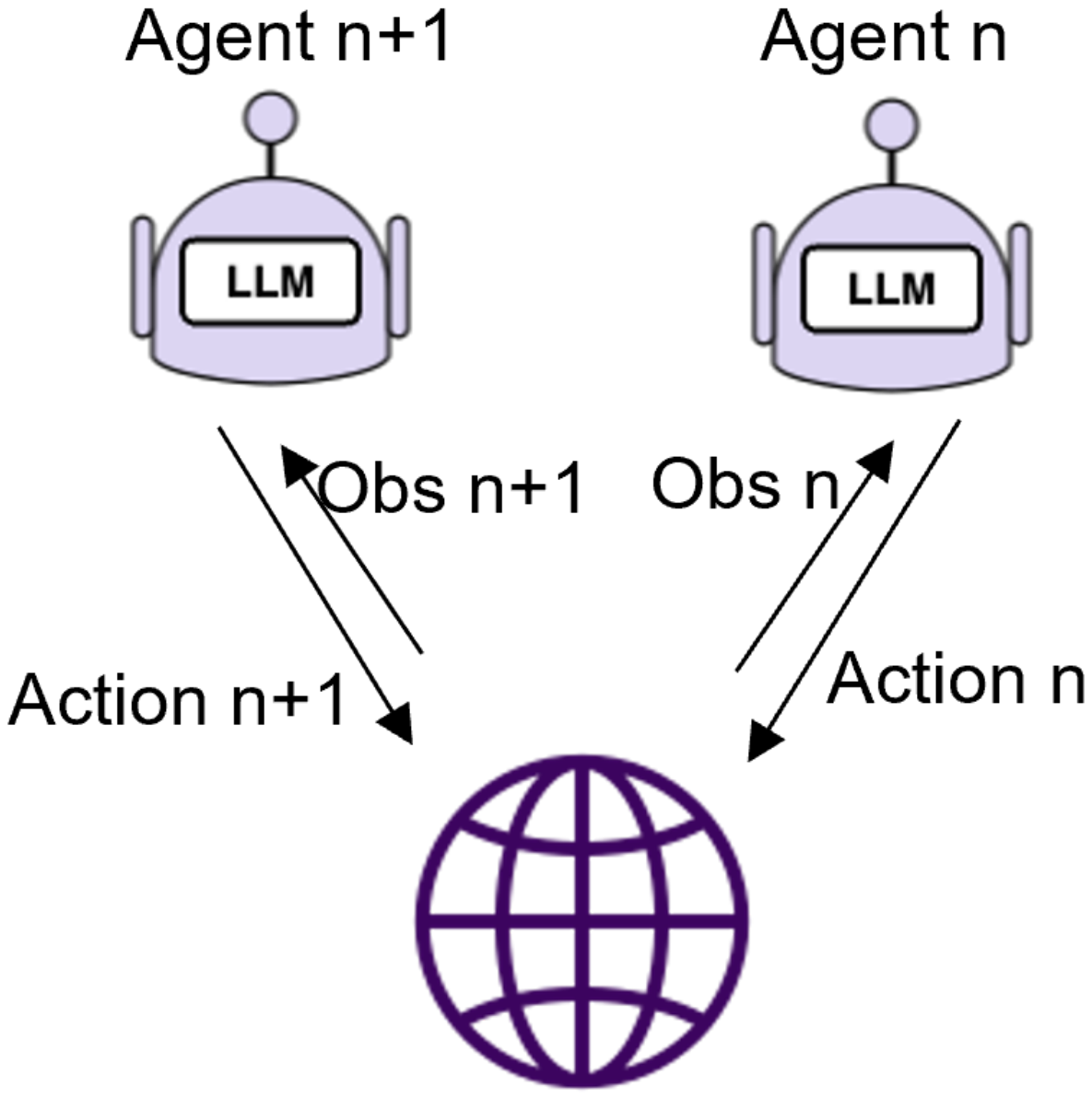}
        \caption{\scriptsize Standalone LLM-powered agents.}
        \label{fig:standalone-agents}
    \end{subfigure}
    \hfill
    \begin{subfigure}[t]{0.3\textwidth}
        \centering
        \includegraphics[trim={0cm 0cm 0cm 0cm}, clip, width=0.7\textwidth]{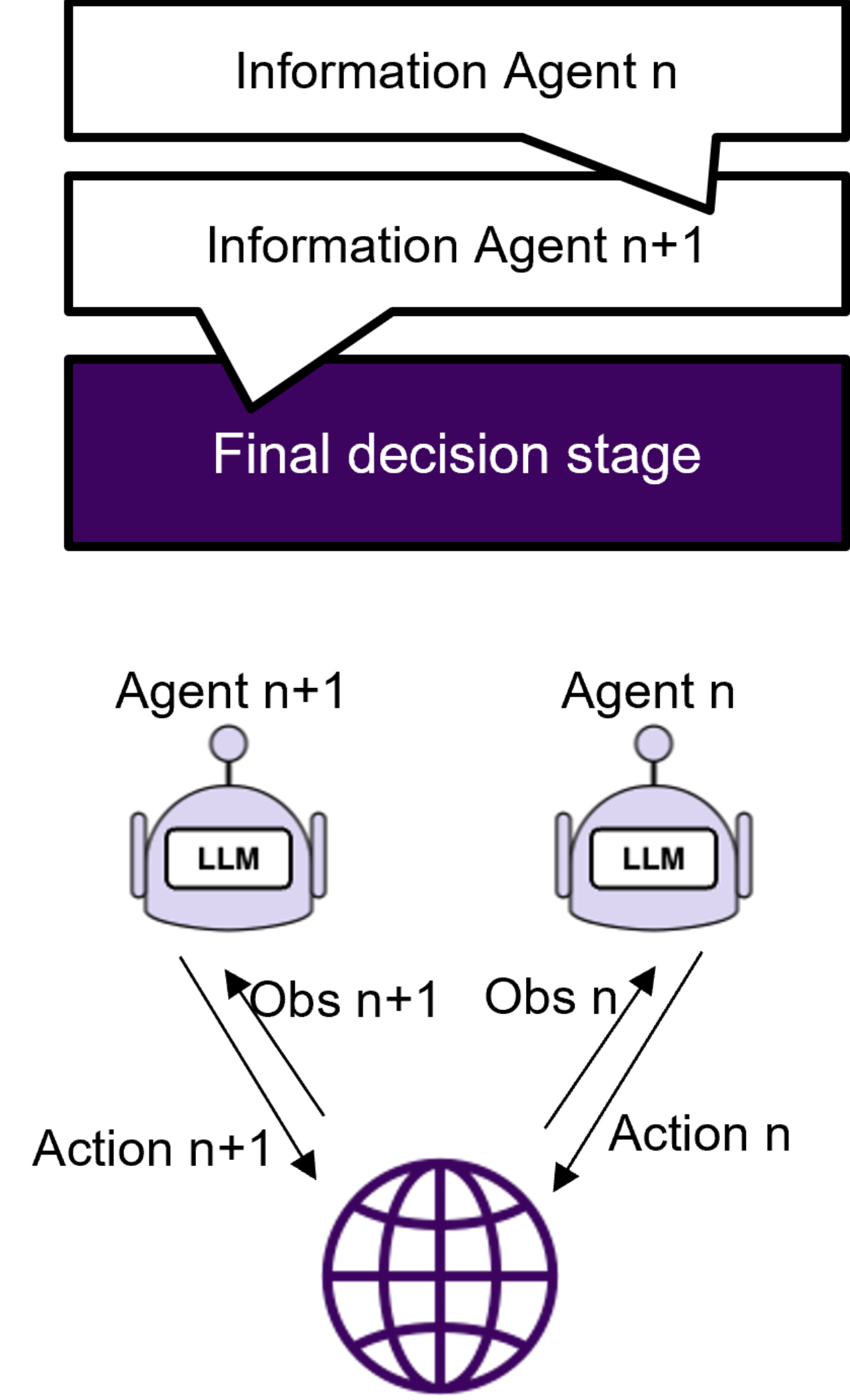}
        \caption{\scriptsize LLM-powered agents with information sharing.}
        \label{fig:info-sharing-agents}
    \end{subfigure}
    \hfill
    \begin{subfigure}[t]{0.3\textwidth}
        \centering
        \includegraphics[trim={0cm 2cm 0cm 0cm}, clip, width=0.7\textwidth]{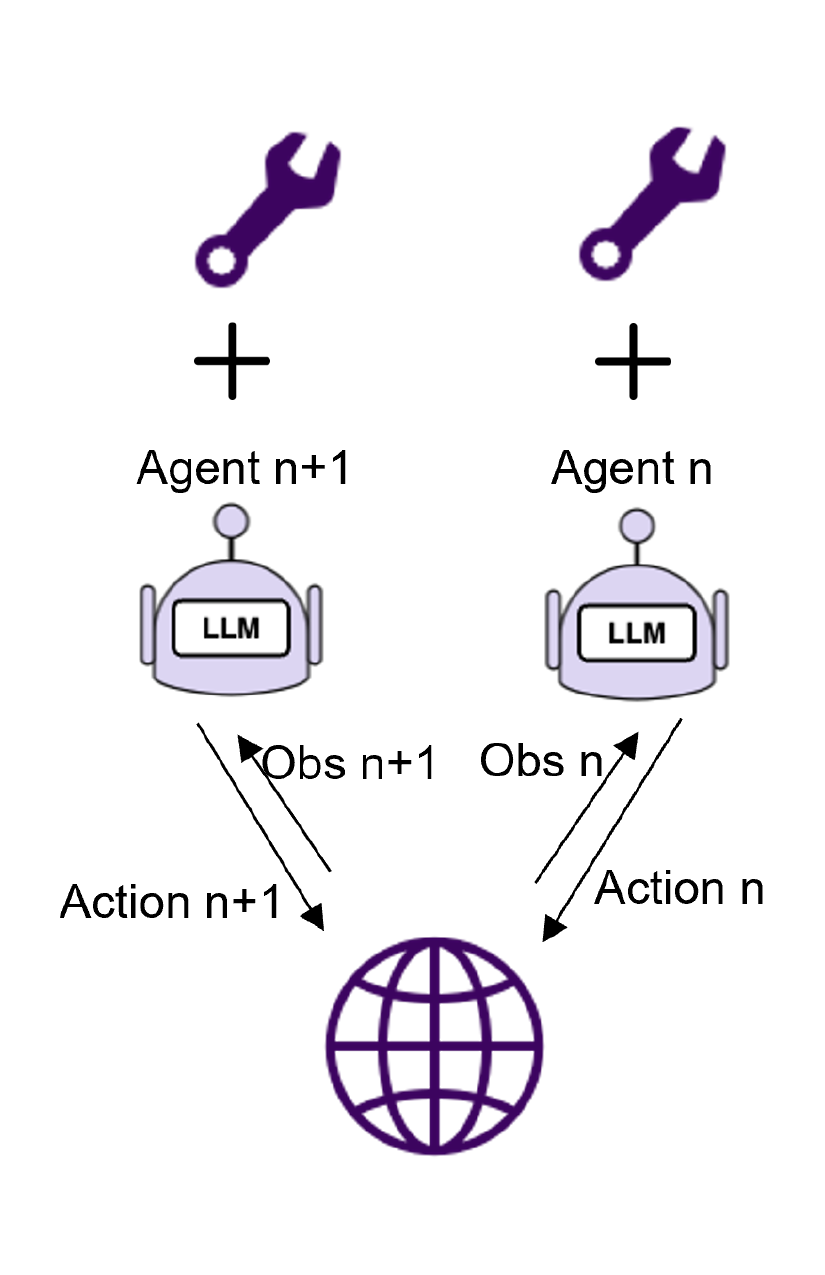}
        \caption{\scriptsize Standalone LLM-powered agents with tool usage.}
        \label{fig:standalone-agents-tool}
    \end{subfigure}
    \hfill
    \begin{subfigure}[t]{0.3\textwidth}
        \centering
        \includegraphics[width=.7\textwidth]{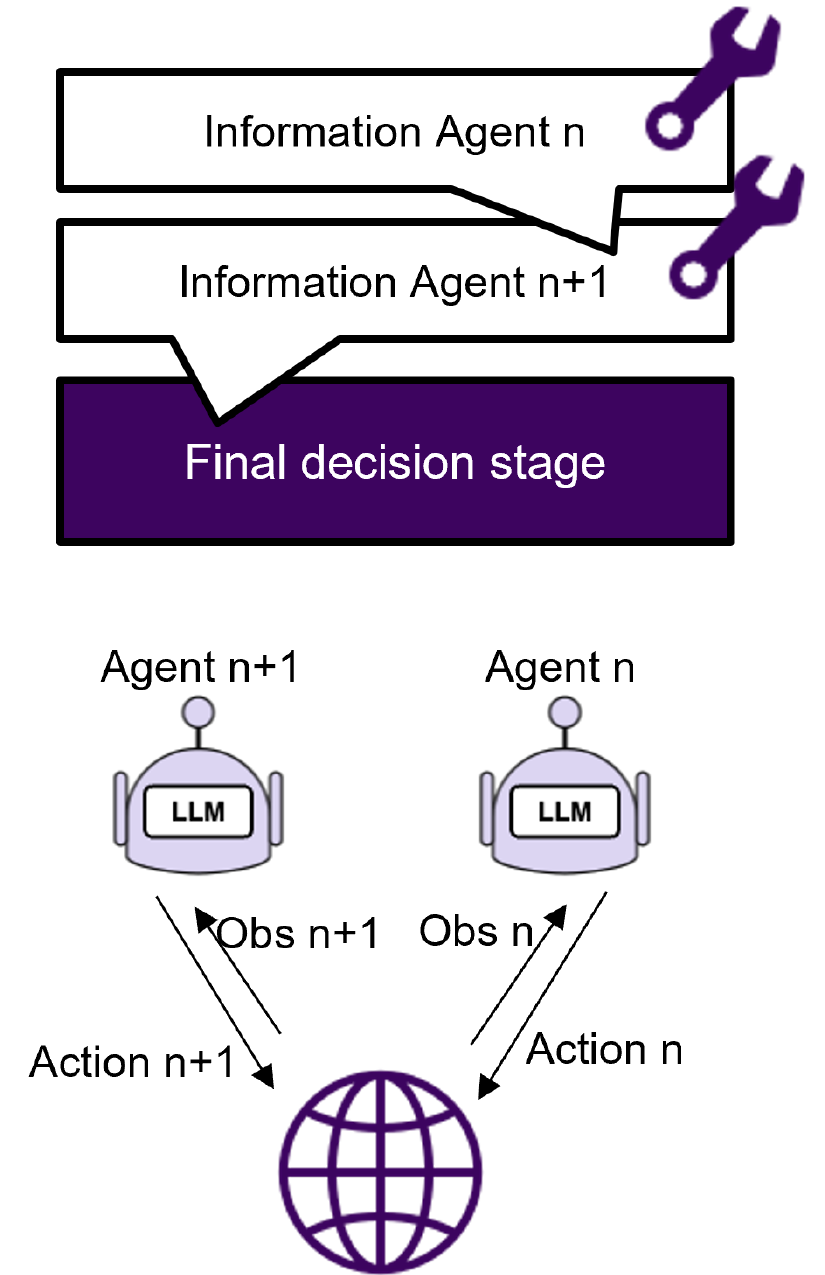}
        \caption{\scriptsize LLM-powered agents with information sharing and tool usage.}
        \label{fig:info-sharing-agents-tool}
    \end{subfigure}%
    ~
    \begin{subfigure}[t]{0.3\textwidth}
        \centering
        \includegraphics[width=.7\textwidth]{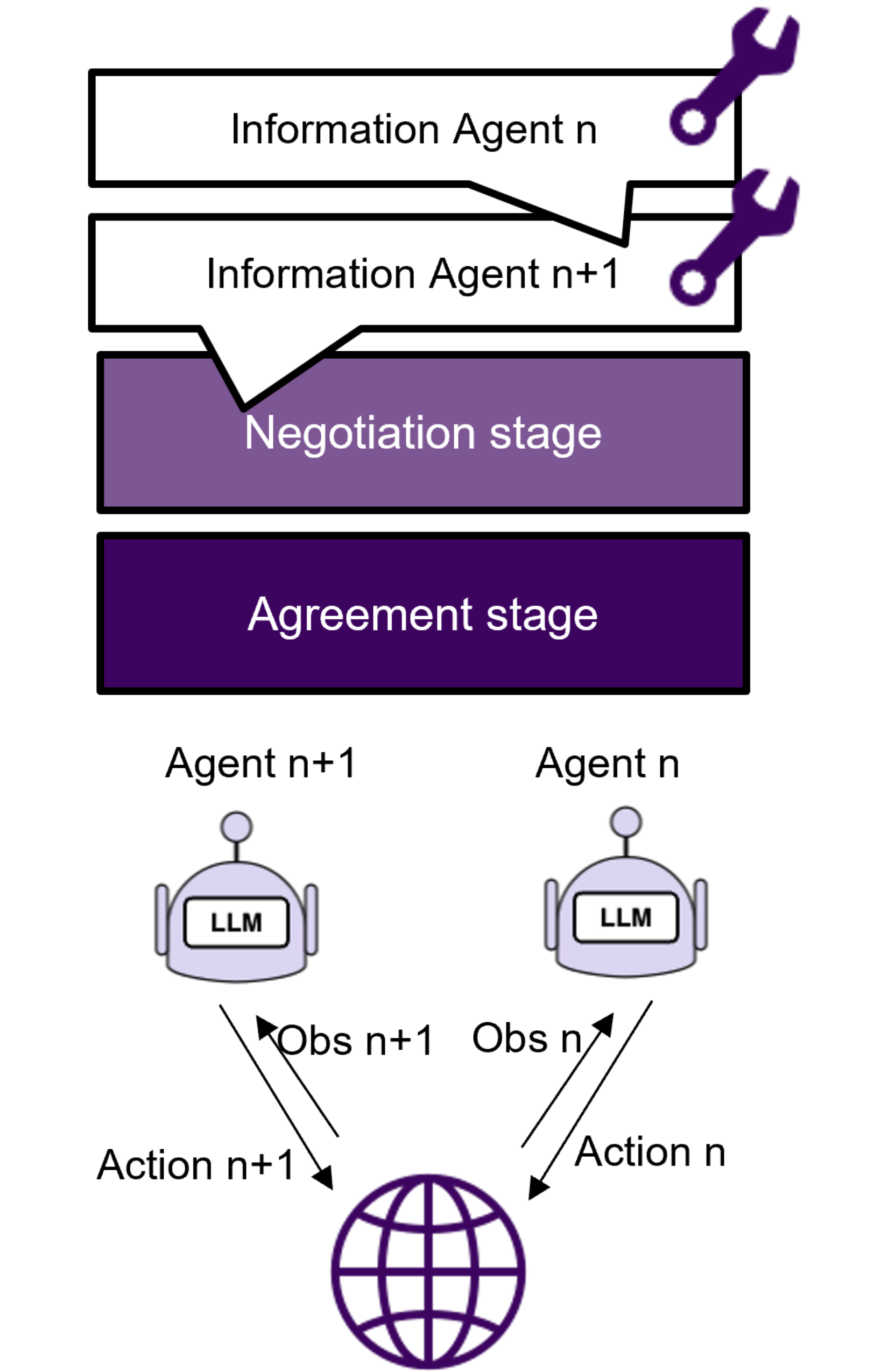}
        \caption{\scriptsize LLM-powered agents with negotiation around tool output.}
        \label{fig:negotiating-agents}
    \end{subfigure}
    
    \caption{LLM-powered consensus-seeking frameworks for the sequential supply chain}
    \label{scalepo}
\end{figure}

An overview of our LLM-powered consensus-seeking frameworks is shown in Fig. \ref{scalepo}. Our frameworks are for "consensus-seeking" because they are orchestrating and steering our SC agents towards an agreement. Whether they ultimately find consensus or not, will depend on the prompt-specific interaction between the agents. In particular, our frameworks include: standalone LLM-powered agents (\ref{fig:standalone-agents}), information sharing between neighbouring agents (\ref{fig:info-sharing-agents}), standalone LLM-powered agents with tools (\ref{fig:standalone-agents-tool}), information sharing between neighbouring agents with tools (\ref{fig:info-sharing-agents-tool}), and negotiation between neighbouring agents (\ref{fig:negotiating-agents}). The underlying inventory management setting for our frameworks comes from the research by \citet{liu-2022}, which we augment by building LLM-powered agents on top of it. 

We utilize existing memory structures to aid the agents in storing previous local observations about the status of the environment, namely the specific agent's inventory state, backlog state and previous orders. Beyond this structure, we rely exclusively on context to create the memory of the agents, also as we progress to consensus-seeking frameworks for decision-making.

\subsection{LLM-powered Decision-Making without Communication}
\subsubsection{Standalone LLM-powered Agents}
\begin{figure}[H]
    \centering
\includegraphics[trim={0cm 2.5cm 0cm 2.5cm}, clip, width=0.9\textwidth]{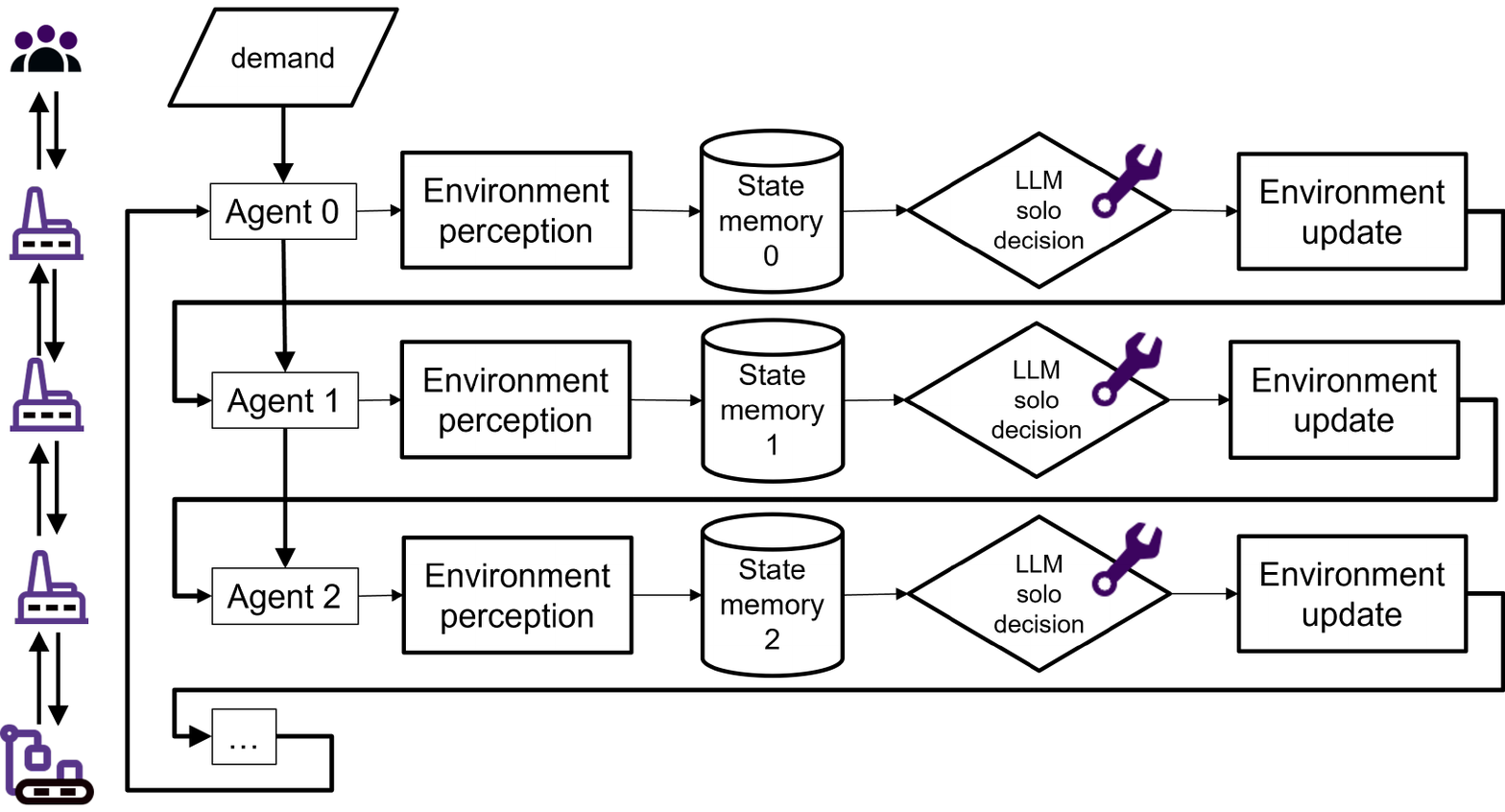}
    \caption{Standalone LLM-powered decision-making process.
    \textbf{Alt Text: Flowchart of a standalone LLM-powered decision making process.} }
    \label{fig:solo-llm-framework}
\end{figure}

The standalone LLM-powered agents constitute the first step in our ablation study. Fig. \ref{fig:solo-llm-framework} illustrates the decision-making process for standalone LLM-powered agents that we employ for a 3-tier sequential supply chain. For each agent, there is the environment perception step, where the agent perceives its observation from the environment. After this, the state is stored in the memory, and the memory is also used to provide some examples for the LLM agent on previous actions. This information from memory and from the observation is contextualized in a prompt for the agent, containing the following elements:
\begin{itemize}
    \item General description of sequential supply chain problem setting. 
    \item Description of objective function:  the function for cost minimization, or the function for the coefficient of variability which we use as a proxy for the bullwhip effect.
    \item Agent’s observation of the environment, containing: inventory, backlog, last order, incoming/transfer orders, demand from downstream neighbour.
    \item Memory: information on previous ten observations on inventory, backlog and order amounts.
    \item Final question on order amount (action) and instruction on formatting LLM output.
\end{itemize}
The result output by the LLM is interpreted as the order amount or action that is executed upon the environment. The input information served to the standalone LLM-powered agent reflects the local view that it has of the supply chain. A detailed breakdown of the prompt components can be found in Appendix \ref{sec:appendix-a}.

\subsubsection{Standalone LLM-powered Agents with Tool Usage}
We introduce the tool on top of the decision made by the standalone LLM-powered agents. The output of the tool embedded in the prompt with the necessary contextualization, to allow the agent to benefit from more information, as illustrated in the example in Appendix \ref{sec:appendix-a}.
The wrench symbol indicates where the tool is invoked in the decision-making process, when the tool is included in the framework. 
\subsection{Communicating LLM-powered Agents}
\begin{figure}[H]
    \centering
    \includegraphics[trim={0cm 3cm 0cm 3cm}, clip, width=0.9\textwidth]{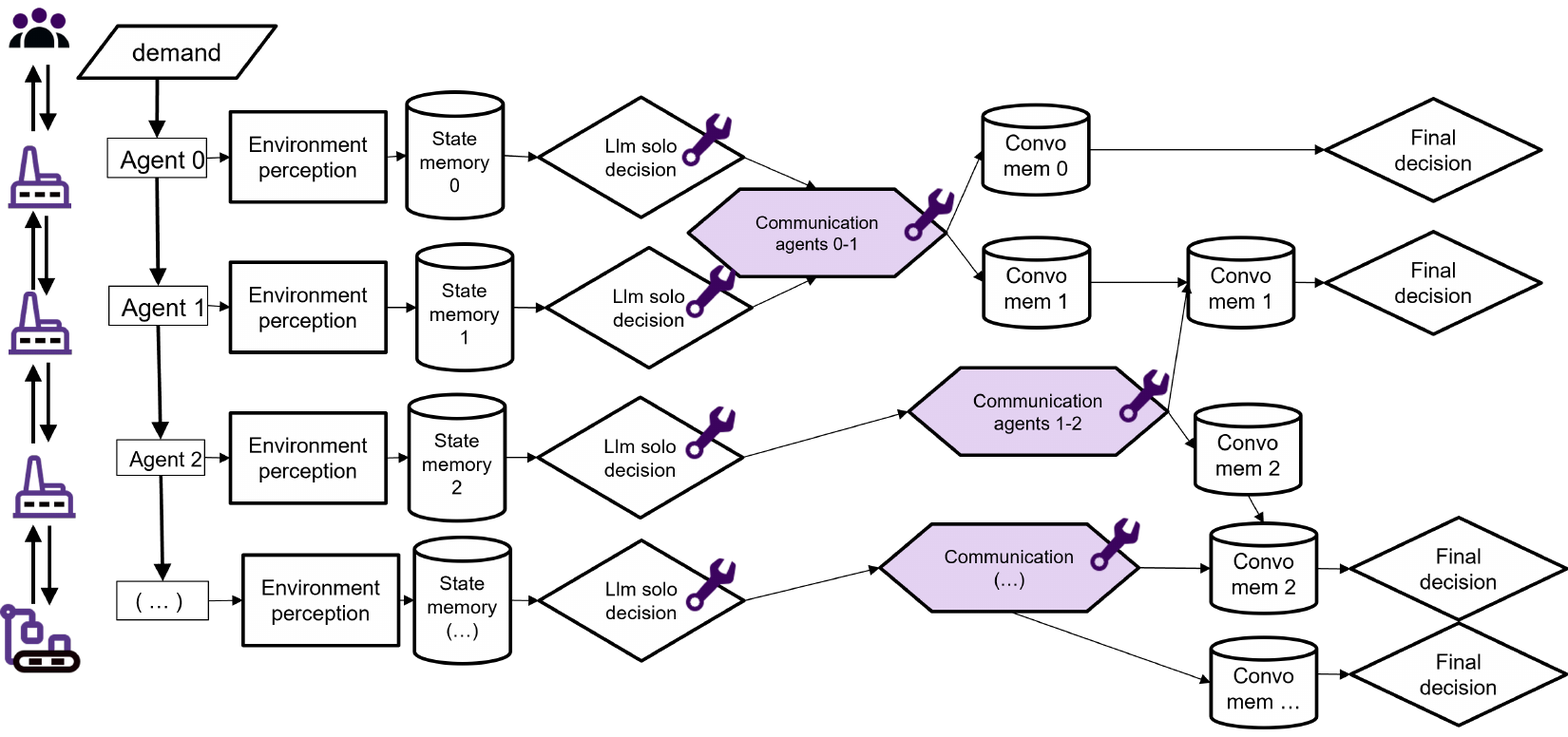}
    \caption{Detailed modular framework for 3-tier supply chain with communication between neighbours.}
    \label{fig:detail-comm-framework}
\end{figure}

Collaboration between agents requires coordination between agents in neighbouring echelons.
We thus develop a cognitive-inspired modular framework that integrates with perception, memory and execution, as illustrated in Fig. \ref{fig:communication-framework} in Appendix \ref{sec:appendix-b} and inspired by the cooperative embodied agents paper by ~\citet{coop-embodied-agents}. At each step of the simulation, after perceiving the environment, each agent can save to and receive information from memory, make a tentative decision based on its standalone viewpoint of the supply chain, and then proceed to a communication stage with a neighbouring agent to seek consensus. Finally, the agent makes a final decision based on all these previous steps and executes it upon the environment. This execution stage after all communication interactions allows the agents to avoid making decisions based on partial information. Fig. \ref{fig:detail-comm-framework} illustrates the details of how information is passed throughout each step for a 3-tier supply chain until it reaches the final decision and execution stage that exercises the final action on the environment.

There are different orchestration frameworks for multi-agent workflows, such as LangGraph ~\citep{langgraph_docs} or AutoGen ~\citep{wu-autogen-convo}. We opted for LangGraph as it offers a simple but versatile implementation and can be easily extended to tool-calling with LangChain. To reflect the partial observability described in ~\citet{sc-coll-hot}, our framework limits communication to neighbouring agents. In particular, our implementation considers a use case where each agent initiates communication with its immediately upstream neighbour. A single interaction therefore always involves two neighbouring agents.

\subsubsection{Communication between Agents} 

There are two types of communication frameworks that we define to be executed between neighbouring agents, corresponding to the stage shaped as a hexagon in Fig. \ref{fig:detail-comm-framework}. These communication frameworks are information sharing and negotiation, which both reflect the partial observability of sequential supply chains by allowing agents to only interact with their immediate neighbours. We begin by defining information sharing, which  involves enriching each agent's information for the standalone decision with information from the neighbouring (upstream) agent. The framework-related choice of implementing interactions with the upstream neighbour comes from the need to mitigate demand amplification, which usually affects upstream echelons in the sequential supply chain; therefore, the framework favours informing upstream agents of details on their downstream neighbours, attempting to steer the former's decision-making process. Intermediate agents in the supply chain still have interactions with both upstream and downstream neighbours before making their final decision. Also this type of communication can be enhanced by including the results from the tool to the information shared between neighbouring agents. 

\begin{figure}[h]
\centering
\includegraphics[trim={0cm 8.5cm 0cm 9cm}, clip, width=0.9\textwidth]{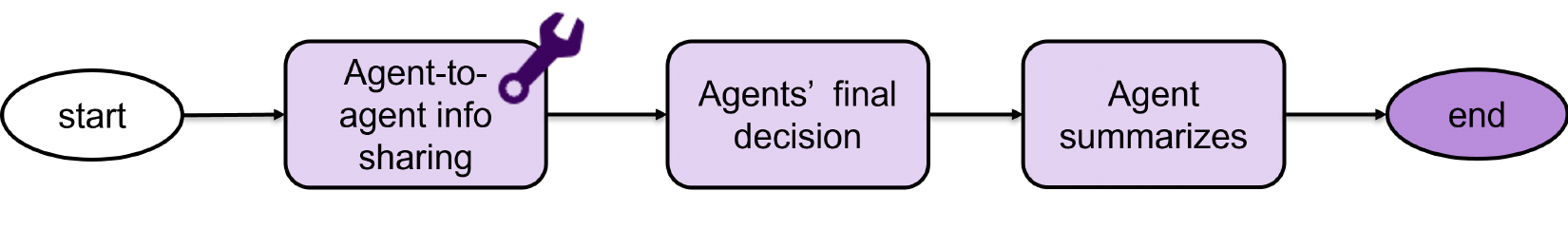}
\caption{Communication framework based on information sharing between neighbouring agents (LangGraph).}

\label{fig:info-sharing-framework}
\end{figure}

Fig. \ref{fig:info-sharing-framework} shows the consensus-seeking framework based on information sharing between neighbouring agents without additional steps taken for negotiation. As we embed the outputs of our tool-calling functions directly in the prompt, the figure applies to both information sharing with tools and without tools. \\

The second type of communication framework, which builds upon the information sharing framework, is the negotiation framework. Negotiation allows for the agents to interact and agree on an order amount, by taking the output of each tool and using these values as upper and lower bounds for a negotiation range. The reasoning behind this comes from the idea of finding a trade-off between the values that each agent would have otherwise decided upon. Therefore, the negotiation framework always includes tool usage, as the tool outputs of communicating agents are used as the starting point for negotiation. The agents have a predefined number of iterations to negotiate the final order amount and subsequently have to name their definitive order amount in a final agreement stage. 

The reasoning behind this is that if the agents manage to find an agreement, this should decrease the amplification of demand as one moves upstream in the supply chain, reducing both costs and the bullwhip effect. For the experiments assessing the global costs of the supply chain, the tool output for each agent constitutes the cost-minimizing amount at each echelon in the sequential supply chain. Our experiments test if a further layer of negotiation on these order amounts improves the global supply chain costs even further.

\begin{figure}[h]
\centering
\includegraphics[trim={0cm 5cm 0cm 5cm}, clip, width=0.9\textwidth]{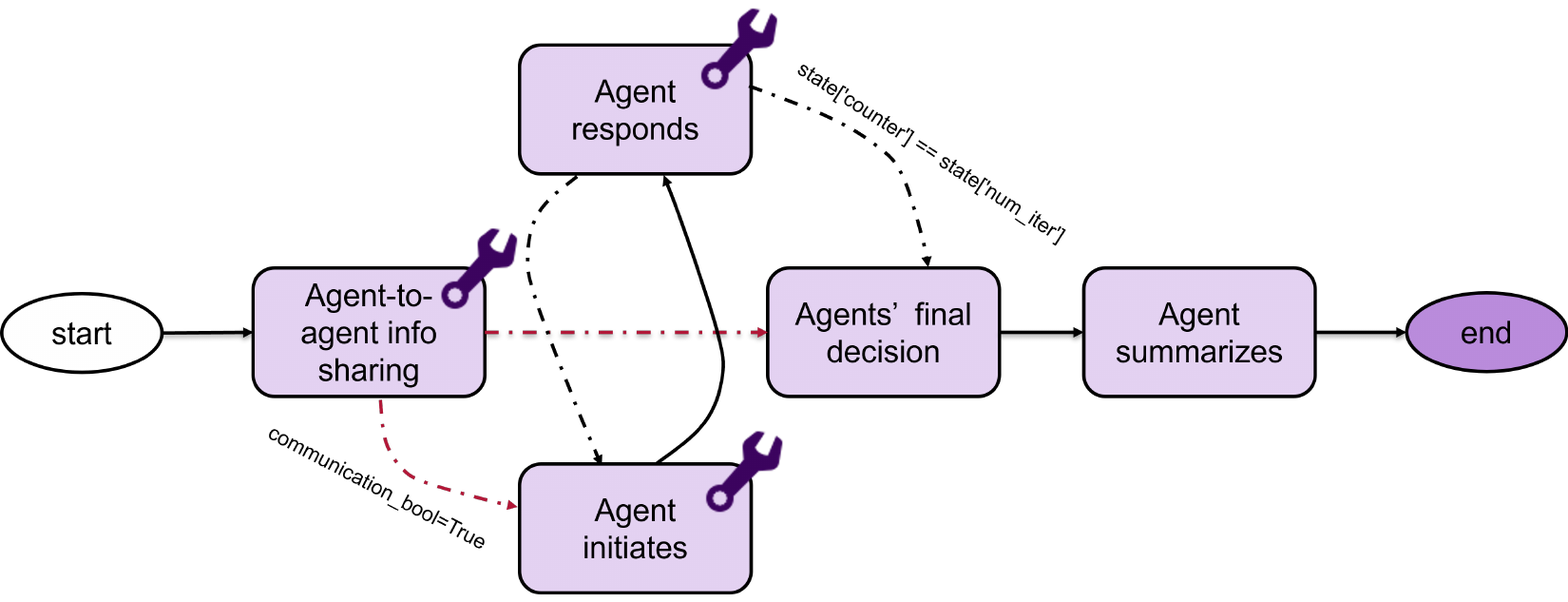}
\caption{Communication framework for negotiation between neighboring agents (LangGraph); for our experiments, state[num\_iter] is equal to 3, meaning there are 3 loops in the communication between neighboring agents.}

\label{fig:langgraph-negotiation-framework}
\end{figure}

The communication framework implemented with LangGraph \citep{langgraph_docs} for each pair of communicating neighbors is illustrated in Fig. \ref{fig:langgraph-negotiation-framework}. Our implementation dynamically creates nodes and edges for a sequential supply chain of any length, without requiring the user to manually define each communication exchange. 

\subsection{Prompt Engineering for different Consensus-Seeking Frameworks and different Metrics}

In this section we discuss the properties of the prompts that we used for different models and metrics. We underline that we undertook manual prompt engineering for our experiments. As a result, the prompts are not fully optimized and largely the same between different foundation models. We will discuss in the Limitations section how model-specific optimization can be achieved in future research.

\subsubsection{Zero-shot Prompting}

Our benchmarking environment utilizes pre-trained LLMs to tackle inventory management challenges. We use zero-shot learning to query our models, thereby requesting that the agents apply their pre-trained knowledge directly to our inventory management tasks, interpreting information that is supply chain-related based on their insights acquired on large knowledge corpora ~\citep{zero-shot}.

We exclude in-context learning from our prompts, because its performance is unstable, and subject to many factors ~\citep{icl-survey}. Additionally, we have opted to keep prompts as similar as possible throughout our experiments, even as we compare the performance with different language models.

Moreover, limiting our prompts to zero-shot learning allows us to minimize the amount of changes we make between different metrics. In particular, the optimization functions change between metrics, as well as the formulation of the question, but the formulation of the observation and description of the sequential supply chain setting remain largely unchanged, as illustrated in Appendix \ref{sec:appendix-a}.

\subsubsection{Prompt Engineering between different Supply Chain Metrics}

Moving from one metric to the other in our implementation only requires adapting the input prompts that get passed into the consensus-seeking frameworks, without changing the implementation of the frameworks themselves and how the messages are passed between agents, showcasing that our implementation is problem-agnostic. In particular, the information sharing and negotiation frameworks apply to multi-agent interactions tackling cost minimization and bullwhip effect mitigation alike. 

\subsubsection{Prompt Engineering to introduce Tool Usage}

As can be observed in Appendix \ref{sec:appendix-a}, the prompts used in our inventory management setting comprise different parts, with details to describe all of the observations and supply chain metrics necessary for each agent to make a decision. As a result, when we want an agent to make a decision based on a tool output, we have to underline the importance of this tool in our prompt. This is achieved by simple means of directives such as inciting the LLM-powered agent through the prompt to give a lot of weight to the tool output in its final decision, as illustrated in Appendix \ref{sec:appendix-x}. The optimal prompts to achieved this results were determined experimentally, and we discuss them in the experimental setup.

\subsection{Tools used by Agents}
We implement two different tools that can be used by the agents based on the metric which they are optimizing. 

 To minimize each agent's costs in the end to end supply chain, we use a demand forecasting tool that uses linear regression to estimate the next order amount, thereby keeping orders similar to past orders and avoiding spikes in demand. The tool utilizes a look-back window of 30 periods to train this model on recent demand data, aiming to anticipate future demand accurately. At the beginning of the simulation, when the dataset is insufficient, the agent defaults to the most recently observed order quantity. 

 To minimize the bullwhip effect of an individual agent in the sequential supply chain, we use a tool that calculates the EOQ formula, as defined in the Problem Setting section, and place the tool output within the LLM prompt.
 
 In both cases, the tool output for the LLM-powered agent is presented to the LLM-powered agent along with a directive on how to use the result in its decision-making.

\section{Experimental Setup}

We run a total of 24 experiments to compare our communication frameworks on different optimization metrics (global bullwhip effect and global costs) and on LLM models of different size. The summary of our experiments is detailed in Table \ref{tab:experiments}. As the temperature used for the LLMs in our experiments is close to zero, the outputs of our runs are near-deterministic. The decision to limit our experiments to specific scenarios without stochastic elements also comes from the fact that running experiments with these models is expensive.

We conduct our experiments with an underlying customer demand that is based on the Merton Jump Diffusion Model, as described in \citet{liu-2022}. We consider an end-to-end supply chain with 3 agents, and a lead time of 2 steps, to reflect the presence of the bullwhip effect as highlighted by \citet{chen-2000} and using the measure from \citet{fransoo-2000} to assess whether the bullwhip effect can be mitigated with our approach. These parameters can be set via the underlying agentic environment from the research by \citet{liu-2022}, which we augment by building on top of it LLM-powered agents.

As we seek to demonstrate improvements in performance (reduction in costs or bullwhip effect) based solely on the sophistication of the consensus-seeking frameworks, the effectiveness of the tool used, and the size of the foundation model powering the agent, we define a fixed maximum order amount for all agent and a common inventory holding cost, backlog cost, variable order cost and fixed order cost for each agent in the sequential supply chain and for each experiment. The fixed parameters across all our agents and experiments are listed in Table \ref{reproduc} in Appendix \ref{sec:appendix-e}.  Consequently, agents/companies only exhibit differing behaviour based on their position in the sequential supply chain. We use the value of 100 as the upper bound for orders across all of the agents in the sequential supply chain and all experiments, which allows increasingly upstream agents to theoretically amplify demand with respect to their downstream agent, starting from the consumer demand, which fluctuates between 0 and 20 units per order. The consumer demand used for our experiments can be found in Appendix \ref{sec:appendix-e}.
For the LLM-powered agents, we enforce this upper threshold with built-in code constraints and explicit prompt engineering to guarantee that the order amount selected by each agent never exceeds 100. This allows us to compare our communication frameworks and baselines in settings that reflect similar possible order quantities.

\subsection{Agent Policies used as Baselines for LLM-powered Agents} 

To compare the performance of our LLM-powered consensus-seeking frameworks with increasing levels of sophistication, we establish different baselines based on traditional extant approaches for restocking. One of the most widely used approaches according to \citet{visentin2021computing} is the (S,s) policy \citep{heur-methods}, therefore we use this as a baseline. The (S, s) restocking policy reorders to a maximum level (S) when inventory falls below a certain threshold (s). In particular, we use values of (S=100,s=60), which reduce the global costs. This decision-making, based on threshold values, can lead to variability amplification up the supply chain, thereby exacerbating the bullwhip effect \citep{chen-2000} and increasing global supply chain costs.

On the other hand, the strong baseline for our experiments is given by the decisions of a tool-based agent: contrary to LLM-powered agents, which embed the tool output into the LLM prompt, these baseline tool agents use the tool output directly as their final decision. Therefore, we have two strong baselines: one for experiments on the global costs metric, and the other for experiments on the global bullwhip effect metric. When optimizing for costs, we obtained a strong baseline by running the tool agent with the output of the linear regression demand forecasting tool, implemented with the previous 30 observations. Similarly, for the bullwhip effect metric, we used the tool agent with EOQ tool, using the previous 30 observations to calculate the average demand from the downstream agent. The ordering costs and holding costs used in the EOQ formula are set to 1 for all agents across all our experiments, as detailed in Table \ref{reproduc}.

\subsection{LLMs used}

We conduct all of our experiments with two different foundation models from the Gemini family, developed by Google Deepmind, namely Gemini Flash and Gemini Pro, to illustrate the performance of our communication frameworks with models of different parameter counts \citep{gemini-models}. Gemini Flash represents a smaller model optimized for fast processing, at the expense of accuracy, while Gemini Pro is larger and is built for more complex, nuanced tasks.  
For both models we run our experiments with a temperature of 0.1. We use a temperature of 0.1 because it is close to 0, which would be almost deterministic; a temperature of 0.1 delivers outputs with a minimum amount of randomness. This is advantageous because it allows for agents to repeat queries which returned an incorrectly-formatted output from the LLM, and deliver results in the correct format when the query is repeated with a small amount of randomness ~\citep{nondet-models}.  

\begin{table}[htbp]
    \centering
    \scriptsize
    \caption{Summary of ablation study of LLM-powered decision making with different metrics and foundation models.}
    \label{tab:experiments}
    
    \begin{tabular}{ccccc}
        $
        \begin{matrix}
        \toprule
        \textbf{Exp. No.} & \textbf{Metric} & \textbf{Tool} & \textbf{Model} & \textbf{Framework} \\
        \midrule
        1   & \text{Global Cost} & - & - & \text{Restocking Policy (S,s)=(100,60)} \\
        2   & \text{Global Cost} & \text{Demand Prediction} & - & \text{Demand Forecasting Tool} \\
        \midrule
        3   & \text{Global Cost} & \text{Demand Prediction} & \text{Gemini 1.5 Flash} & \text{Standalone LLM} \\
        4 & \text{Global Cost} & \text{Demand Prediction} & \text{Gemini 1.5 Flash} & \text{LLM with Info Sharing} \\
        5   & \text{Global Cost} & \text{Demand Prediction} & \text{Gemini 1.5 Flash} & \text{Standalone LLM + Tool} \\
        6 & \text{Global Cost} & \text{Demand Prediction} & \text{Gemini 1.5 Flash} & \text{Info Sharing + Tool} \\
        7  & \text{Global Cost} & \text{Demand Prediction} & \text{Gemini 1.5 Flash} & \text{Negotiation + Tool} \\
        \addlinespace
        8   & \text{Global Cost} & \text{Demand Prediction} & \text{Gemini 1.5 Pro} & \text{Standalone LLM} \\
        9   & \text{Global Cost} & \text{Demand Prediction} & \text{Gemini 1.5 Pro} & \text{LLM with Info Sharing} \\
        10   & \text{Global Cost} & \text{Demand Prediction} & \text{Gemini 1.5 Pro} & \text{Standalone LLM + Tool} \\
        11   & \text{Global Cost} & \text{Demand Prediction} & \text{Gemini 1.5 Pro} & \text{Info Sharing + Tool} \\
        12   & \text{Global Cost} & \text{Demand Prediction} & \text{Gemini 1.5 Pro} & \text{LLM Negotiation + Tool} \\
        \midrule
        13 & \text{Global Bullwhip} & - & - & \text{Restocking Policy (S,s)=(100,60)} \\
        14   & \text{Global Bullwhip} & \text{EOQ} & - & \text{EOQ tool} \\
        \midrule
        
        15 & \text{Global Bullwhip} & \text{EOQ} & \text{Gemini 1.5 Flash} & \text{Standalone LLM} \\
        16 & \text{Global Bullwhip} & \text{EOQ} & \text{Gemini 1.5 Flash} & \text{LLM with Info Sharing} \\
        17 & \text{Global Bullwhip} & \text{EOQ} & \text{Gemini 1.5 Flash} & \text{Standalone LLM + Tool} \\
        18  & \text{Global Bullwhip} & \text{EOQ} & \text{Gemini 1.5 Flash} & \text{Info Sharing + Tool} \\
        19  & \text{Global Bullwhip} & \text{EOQ} & \text{Gemini 1.5 Flash} & \text{Negotiation + Tool} \\
        \addlinespace
        20  & \text{Global Bullwhip} & \text{EOQ} & \text{Gemini 1.5 Pro} & \text{Standalone LLM} \\
        21 & \text{Global Bullwhip} & \text{EOQ} & \text{Gemini 1.5 Pro} & \text{LLM with Info Sharing} \\
        22  & \text{Global Bullwhip} & \text{EOQ} & \text{Gemini 1.5 Pro} & \text{Standalone LLM + Tool} \\
        23 & \text{Global Bullwhip} & \text{EOQ} & \text{Gemini 1.5 Pro} & \text{Info Sharing + Tool} \\
        24 & \text{Global Bullwhip} & \text{EOQ} & \text{Gemini 1.5 Pro} & \text{Negotiation + Tool} \\
        \bottomrule
        \end{matrix}
        $
    \end{tabular}
    
\end{table}

 \section{Results \& Discussion}

To illustrate the performance of our consensus-seeking communication frameworks, we conduct experiments focusing on two different problem settings for inventory management in the sequential supply chain: global cost minimization and global bullwhip effect minimization.

\subsection{Global Cost Minimization}
\begin{figure}[H]
\centering 
\includegraphics[scale = 0.5]{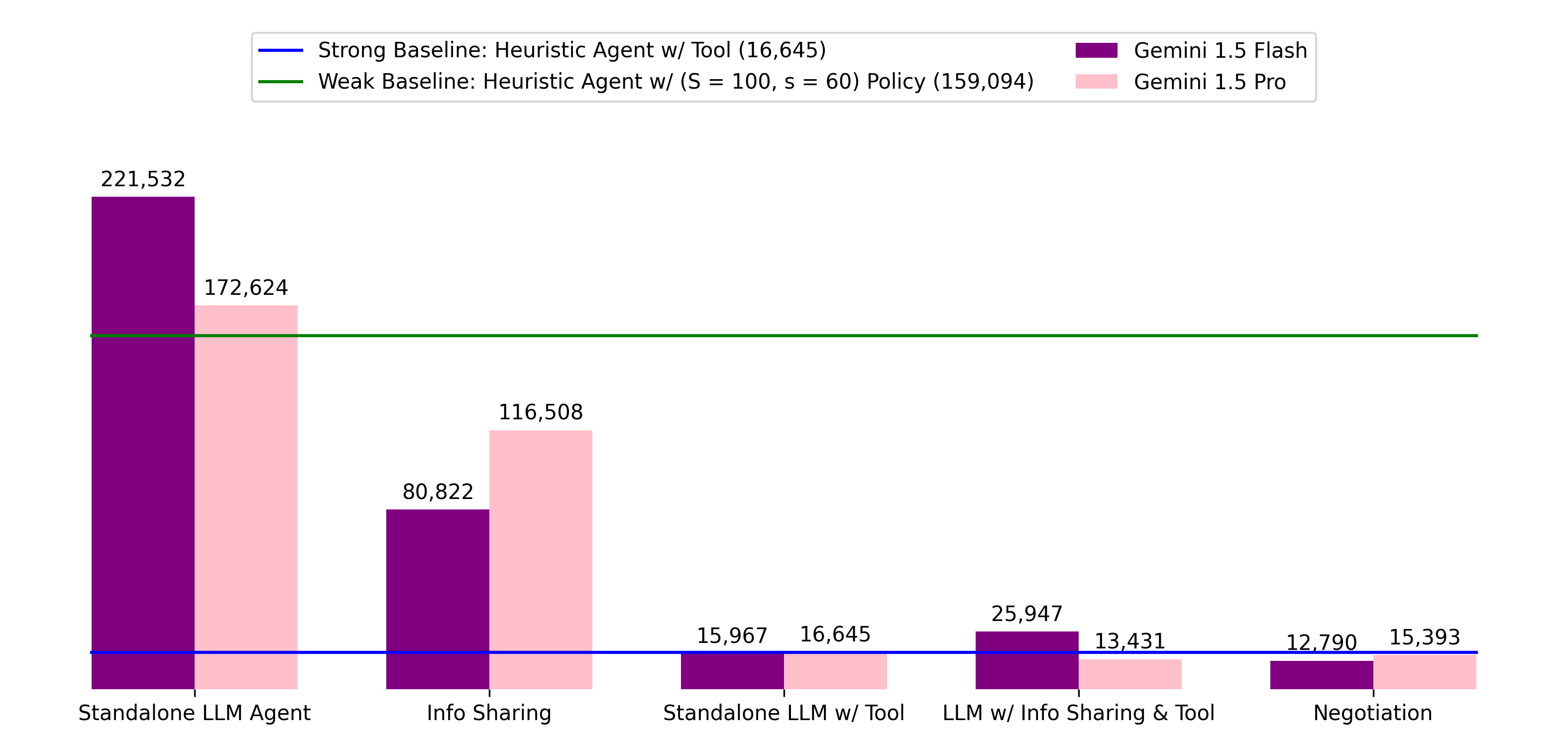}
\caption{Global cumulative costs achieved by different consensus-seeking frameworks.}

\label{fig:global-costs-perf}
\end{figure}

Fig. \ref{fig:global-costs-perf} visualizes the global costs achieved by each consensus-seeking framework, for two foundation models, and compares the results to the two baselines achieved by the baseline agent implementations: a weak baseline given by the performance of an agent employing an (S=100,s=60) restocking policy, and a strong baseline given by the performance of an agent using a demand forecasting tool achieved with linear regression. 

\subsubsection{Cost Performance of Gemini Flash}
For Gemini Flash, we observe a gradual cost reduction as we introduce the demand forecasting tools, and as we introduce increasing levels of sophistication in the consensus-seeking framework, first with information sharing between neighbouring agents, and subsequently with negotiation around the value of the tool output. In particular, for Gemini Flash, we observe:
\begin{itemize}
    \item  63.5\% cost reduction when adding tools to the standalone agents,
    \item  92.8\% cost reduction when comparing information sharing agents to standalone agents,
    \item  67.9\% cost reduction when adding tools to information sharing agents,
    \item  50.7\% cost reduction when adding negotiation to agents implementing information sharing with tool usage. 
\end{itemize}  
This is in line with our expectations that communication between neighbouring agents and additional information about the status of the supply chain can aid individual agents in making more informed decisions. Additionally, the only framework that underperforms the weak baseline is the standalone LLM agent without tool usage.
\subsubsection{Cost Performance of Gemini Pro}
For Gemini Pro, we observe that the trend in cost reduction is similar to that obtained with Gemini Flash, but there is not a consistent performance improvement achieved when using a larger model, even if both models are from the same developer organization (Google DeepMind). Existing research confirms that prompt-based interactions are still brittle ~\citep{10.1145/3544548.3581388} and that ``performance can vary non-monotonically with model size'' ~\citep{emergent-abilities-llms}. \citet{prompt-eng-consistency} also finds that different prompts had variable effects across various models. Our case study optimizes the prompts for the smaller model, Gemini Flash, using manual optimization techniques. For the Gemini Pro, the prompts are identical except for the way the tool output is introduced, as can be observed in Fig. \ref{fig:tool-promps} in Appendix \ref{sec:appendix-x}. Prior to these prompt modifications, the performance achieved with a larger model was more unstable. Table \ref{cum-glob-cost} in Appendix \ref{sec:appendix-c} shows the performance of our consensus-seeking frameworks relative to global costs, considering a Merton Jump Diffusion (spike) demand and a prediction tool for demand forecasting.

Finally, it is also interesting to note that, for both models used, the performance of the negotiation framework beats the hard baseline involving tool-based restocking policy by the agent. This shows that LLM-powered agent interactions with high degrees of orchestration can outperform even mathematical tool-driven performances.

\subsubsection{Effect of Metric Complexity on Tool Performance}
The experiments relative to the cost minimization objective illustrate that, when a tool is highly suited for a given metric, such as our linear regression tool to predict future demand and minimize costs, the highest performance is achieved by those frameworks that give strong weight to the tool itself, without necessarily needing an elaborate communication framework between agents. Despite this, additional orchestration on top of the tool implementation, such as with the negotiation framework, can improve tool-driven results even further. This is in line with existing SC research that highlights how sophisticated tools in the form of algorithms can yield optimization of resource allocation and cost reductions \citep{bu2024logistic}. 

However, demand forecasting to minimize costs is a relatively simple challenge in inventory management, defined by a formula which considers a simple sum of different costs. This is simple enough for a foundation model, which is a probabilistic machine, not a tool suited for complex calculations. In cases where the metric is more complex, such as the minimization of the bullwhip effect via a coefficient of variability, communication between neighbouring agents to increase awareness about the status of the supply chain becomes crucial. In these cases, the communication frameworks become the key factors in improving performance, as we illustrate in the following results. 

\subsection{Global Bullwhip Effect Minimization}

Table \ref{fig:bullwhip-perf} shows whether or not the global bullwhip effect achieved by each consensus-seeking framework is below 1. A value below 1 is desirable as it indicates a negligible amount of bullwhip effect, according to the metric by \citep{fransoo-2000} which we used in our inventory management setting and is already implemented by \citet{liu-2022}. Table \ref{fig:bullwhip-perf} illustrates the results of our consensus-seeking frameworks, as well as the results achieved by the baseline agents. In particular, the strong baseline, based on the agent that calculates the EOQ and uses this amount as the order quantity, successfully reduces the bullwhip effect to a value below 1. 

Regarding our LLM-powered consensus-seeking frameworks, we can observe that introducing a tool to the standalone LLM-powered agent does not sufficiently mitigate the bullwhip effect and, in the case of Gemini Flash, even worsens the results with respect to the bullwhip effect achieved by standalone agents without tool. Performance augmentations become apparent with the introduction of communication-based frameworks: information sharing, information sharing with tool, and negotiation.

Importantly, as can be observed in Table \ref{bullwhip-res} in Appendix \ref{sec:appendix-c}, the best bullwhip effect performance was achieved with the negotiation framework. In particular, the negotiation framework yields a 66.2\% bullwhip effect reduction compared to information sharing with tool in the case of Gemini Pro, and a 33.2\% bullwhip effect reduction in the case of Gemini Flash. This significant jump is the result of neighbouring agents attempting to agree on order amounts, thereby reducing demand amplification \citep{optim-methods-bw}. 

\subsubsection{Detailed Example of Negotiation for Bullwhip Effect Mitigation}
\begin{figure}[H]
\centering
\includegraphics[trim={0cm 2.5cm 0cm 2.5cm}, clip, width=0.9\textwidth]{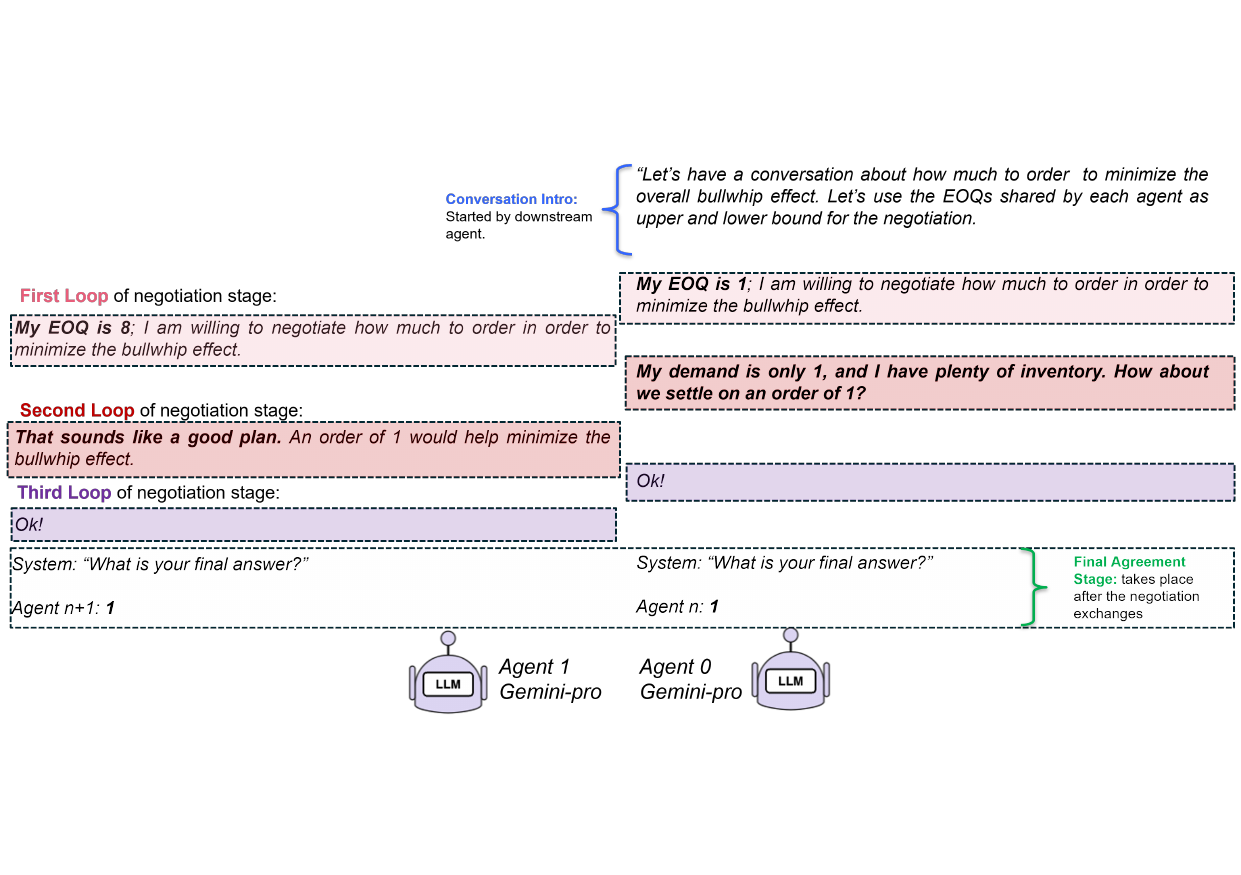}
\caption{Negotiation for Bullwhip Effect – Detailed Example.}

\label{example-negotiation-1}
\end{figure}

\begin{figure}[H]
\centering
\includegraphics[trim={0cm 4.5cm 0cm 4.5cm}, clip, width=0.9\textwidth]{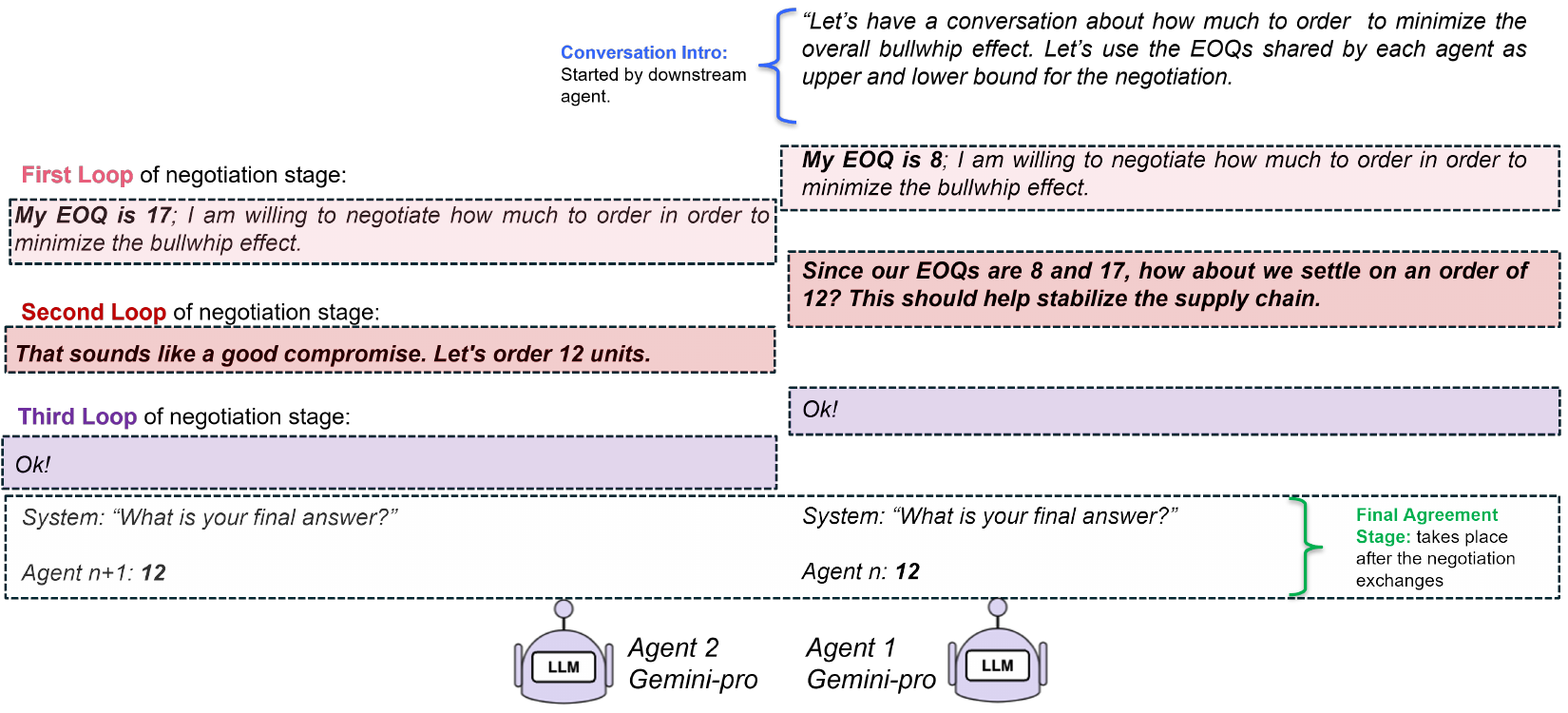}
\caption{Negotiation for Bullwhip Effect – Detailed Example cont.}

\label{example-negotiation-2}
\end{figure}

\begin{figure}[H]
\centering
\includegraphics[trim={0cm 4.5cm 0cm 4.5cm}, clip, width=0.9\textwidth]{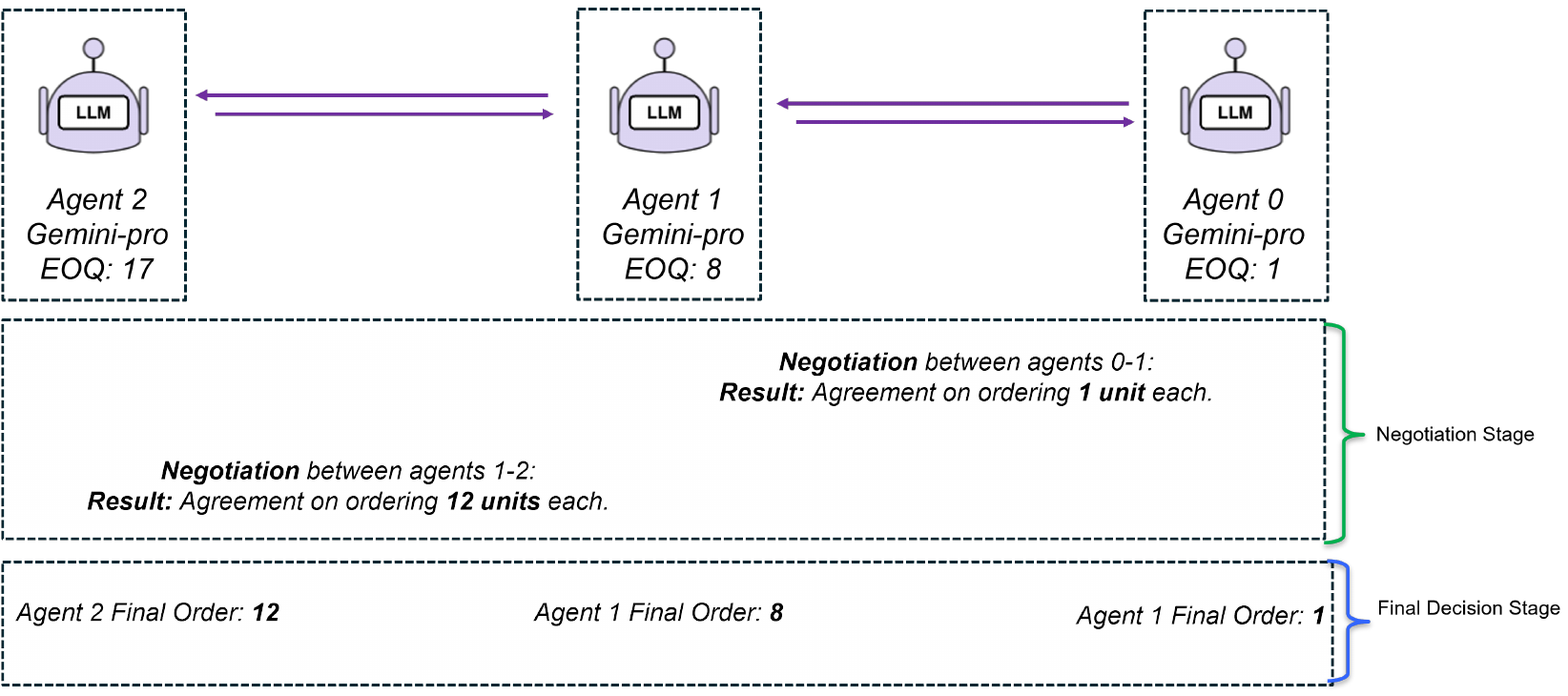}
\caption{Negotiation for Bullwhip Effect – Detailed Example cont.}

\label{example-negotiation-3}
\end{figure}

In Figures \ref{example-negotiation-1}, \ref{example-negotiation-2} and \ref{example-negotiation-3}, we illustrate an example of interactions between neighbouring agents across our sequential supply chain. From Figures \ref{example-negotiation-1} and \ref{example-negotiation-2} we observe that agents are able to use information acquired previously from observing the environment to motivate their desired order amounts and convince their neighbouring upstream agent. The are also able to propose trade-offs and agree on them. Appendix \ref{sec:appendix-d} illustrates some further examples from our experiments of agents' agreements and occasional disagreements in negotiation.

From these examples, two observations are clear. Firstly, the behavior of LLM-powered agents depends on the information shared previously between agents, steering the agents to propose a certain order amount to their neighbor. Secondly, the prompts used to define the setting, as well as the choice of the agent used to deliver this message, have a large effect on the resulting agent behavior. 

Our experiments reveal that when not explicitly directed on which strategy should be adopted for selecting ordering amounts, LLM-powered agents primarily use the average strategy although they occasionally use some other strategies, such as going for one extreme of the negotiation interval or disagreeing altogether. This is in line with research on LLM-powered MAS and their consensus-seeking strategies \citep{mas-survey}. To use the same terminology from these authors, our LLM agents in negotiation tend to be either ``stubborn'' or ``suggestible''. Stubborn agents tend to dominate the final decision, whereas suggestible agents tend to behave more collaboratively. In the case of our specific supply chain setting, the downstream agent initiates the conversation and explains the nature of the negotiation, and subsequently tends to assume a more stubborn attitude. The upstream agent responds and in most cases tends to assume a more suggestible attitude. This particular behavior is appealing for a use case that aims to minimize demand amplification, as the insights from downstream prove to benefit this optimization, but may have to be modified in other use cases.

\subsubsection{Framework Comparison for Bullwhip Effect Performance}
As discussed in the Problem Setting section, introducing negotiation as a way for agents to agree on order amounts starting from the EOQ should mitigate the bullwhip effect for the global supply chain \citep{optim-methods-bw}. This is clearly observable from our results: introducing more sophisticated consensus-seeking frameworks (information sharing, information sharing with tool, and negotiation) yields a bullwhip effect metric below 1 for both foundation models. In particular, for both models, the lowest (best) bullwhip effect measures were achieved with the negotiation framework. This result experimentally backs the research suggesting that information sharing and agreeing on order amounts improves the bullwhip effect \citep{bullwhip-progress-trends, bullwhip-info-enrichment}.

Table \ref{bullwhip-res} in Appendix \ref{sec:appendix-c} shows the experimental results relating to the bullwhip effect for a sequential supply chain of 3 echelons given different LLM-powered consensus-seeking frameworks and baselines.

Throughout our experiments, we observe that end-to-end supply chains can operate much better compared to extant restocking policies when adopting LLM-powered settings that include coordinated communication and tool usage. Additionally, when LLM-powered agents are handled within a negotiation framework, their behavior converges to best practices in SC literature to lessen the bullwhip effect.

\begin{table}[H]
\caption{Global bullwhip effect resulting from different consensus-seeking frameworks.}
\resizebox{\textwidth}{!}{%
\begin{tabular}{lccccc} 
$
\begin{matrix}
\toprule
\multicolumn{6}{c}{\text{Criteria for negligible amount of bullwhip effect\textsuperscript{a}: }   < 1} \\
\scriptsize
\midrule
\text{Weak baseline} & \text{\xmark}  \\
\text{Strong baseline} & \text{\cmark}  \\
\midrule
\text{Decision Framework} & \text{Standalone LLM} & \text{Standalone LLM + tool} & \text{Info Sharing} & \text{Info Sharing + tool} & \text{Negotiation} \\
\midrule

\text{smaller model (Gemini Flash)} & \text{\cmark} & \text{\xmark} & \cellcolor{blue!25}\text{\cmark}  & \cellcolor{blue!25}\text{\cmark} & \cellcolor{blue!25}\text{\cmark}  \\
\text{larger model (Gemini Pro)} & \text{\xmark} & \text{\xmark} & \cellcolor{blue!25}\text{\cmark} & \cellcolor{blue!25}\text{\cmark} & \cellcolor{blue!25}\text{\cmark} \\
\bottomrule
\end{matrix}
$
\end{tabular}
}
\tabnote{\textsuperscript{a}A value below 1 (indicating the coefficient of variability of demand is below 1) indicates that the bullwhip effect is negligible.}
\label{fig:bullwhip-perf}
\end{table}

\section{Managerial Implications and Implementation}

Reaching consensus on operational decisions across an end-to-end supply chain is time-consuming and often not possible due to the amount of manual effort involved. In the literature, automation of consensus-seeking in low level supply chain decisions using agent based systems were proposed since a long time, but not adopted by industry. Previously proposed AI-powered agents in SCM were challenging to implement for companies due to the skill set needed for their development, the complexity of finding training data and the need for all supply chain agents in a given ecosystem to match their input/output formats. In particular SMEs cannot afford large data science teams required to implement complex artificial intelligence approaches and manage large databases. Recent advances in pre-trained LLMs offer an opportunity: the possibility to work with a fast, natural-language based interface, that has the ability to perform industrial control tasks \cite{song-industrial-control} without the need to set up interoperability across multiple supply chain information systems. Recent advances in the use of LLM-agents for consensus-seeking also prompted us to explore whether they can be used in a supply chain setting \citep{tessler2024ai}.

In this paper, we developed and provided a suite of LLM-powered consensus-seeking frameworks for  supply chains and tested them with a Bullwhip effect simulator. Our experiments yield interesting managerial implications, which we report across four categories: performance, scalability, implementation, and reliability.

\subsection{Reliability and Explainability}

Our experiments showed that neither foundation model size and nor tool reliability are a guarantee to achieve high performance for global cost mitigation and global bullwhip effect mitigation. Larger models do not yet yield significant performance gains over smaller models in this setting, and each model that is plugged into our consensus-seeking frameworks requires a high engagement from a supply chain specialist or engineer to optimize the prompts and yield satisfying results. This implies that we are still far from autonomous LLM-powered supply chains and decisions made to optimize real-world supply chain metrics still require a human-in-the-loop approach. 

In addition to this, using tools is not a guarantee for success. In particular, success in tool-driven frameworks depends on the the tool and the prompting skills of an engineer in guiding the foundation model on how to best use the tool output in the LLM prompt. The LLM thus acts similar to a human agent that needs input on how to work with a tool - but with the added benefit of being a scalable automated solution.


In our experiments relating to the bullwhip effect, the EOQ tool implementation is less performant at lower levels of the framework (standalone implementation with tool): performance improvements are achieved with communication, in particular through negotiation. This is in line with supply chain research suggesting agreement on order amounts mitigates the bullwhip effect \citep{prompt-eng-consistency}. Managers should be wary of placing too much trust in LLM-powered decision-making, as the outputs they provide are only as strong as the efforts by engineers to curate meaningful prompts and guide the consensus-seeking process through effective agent orchestration. 


\subsection{Performance}
Experiments relating to global cost minimization highlight that tool-usage is only effective when the function behind the tool is actually effective in tackling a certain challenge. Relying on standalone LLM-agents to not outperform existing solutions, but adding a tool does lead to performance improvement. For example, in the case of the linear regression model used for demand forecasting, this tool keeps the demand similar to previous observations, preventing the demand spikes that cause costs to soar. 


Another significant advantage of our consensus framework is that it is problem-agnostic. Agent interactions are created dynamically, based on the type of interaction (information sharing vs negotiation, etc.) as decided by the user. Such an approach can scale to supply chains of arbitrary length, and be easily adapted to guide interactions relative to different supply chain challenges. In particular, the structure and sequence of negotiations - starting with neighbouring agents  on a one-on-one basis and subsequently making decisions across the entire sequence of agents, reflect how company-to-company interactions take place in real-world supply chains \citep{zhu2024effects}.

It is also interesting to point out that the results of our experiments outperform the hard baselines (tool-based) in many cases, even if the prompts used in our experiments have been optimized manually and not optimized to fit a specific LLM. This result is encouraging from a managerial perspective, because it shows that good results can be achieved also when prompts are engineered by a supply chain expert who is not necessarily a prompt engineering specialist. 

\subsection{Scalability}

Many real world supply chain consensus problems require settings where multiple agents and multiple objectives can coexist. Therefore scalability of the communication framework is an important consideration. The complexity in our  framework for a sequential supply chain is $O(n)$, whereas the complexity in a supply chain network with $m$ layers and $n$ echelons would be $O(m \cdot n)$. Preserving efficiency could be achieved in future work by using quick and safe algorithms to identify which agents have high impact on a specific outcome and subsequently limiting communication to those high-impact agents, as proposed by \citet{schoepf-identifying}.


\subsection{Implementation} 

One of the key advantages of using LLM-powered agents for consensus-seeking is that they do not require interoperability across the supply chain information ecosystem for applications to work effectively. In particular, some companies will move towards LLM-powered decision-making before others, but this will not prevent them from engaging with companies that still make decisions with extant restocking approaches or hybrid solutions. Since LLM-powered agents receive inputs in natural language and output results in a human-understandable format, they can coexist with other technologies used by other agents, be it human or AI, in the supply chain. This feature is  appealing for real-world use cases and could be investigated further with our communication framework, by exploring scenarios where agents use different tools or cognitive models to make decisions about inventory management challenges. 

Moreover, as the trend toward expanding context windows (i.e., maximum input lengths) in LLMs continues, foundation models will process increasingly extensive sequences of data \citep{gartenberg-2024}, allowing them to analyse complex supply chain scenarios. With the advent of Industry 5.0, the significance of General AI is set to deepen even further \citep{ivanov2023industry}. As AI in SCs becomes more widespread, integrating LLM-powered consensus-seeking frameworks in existing digital twin structures will allow company-tailored information to guide the agent's decision-making process \citep{jackson2024generative}, for which real-time access to information in the whole company is essential.


\section{Limitations}

As a technology with limited benchmarking environments in the supply chain management community, LLM-powered agents are still at a very early stage in reaching their potential. A general weakness of foundation models, independently of the rules and frameworks that surround them, is that the performance when switching between different foundation models is still unstable. This is in line with computer science literature on the topic, for example \citet{prompt-eng-consistency}. This limitation of our environment could easily be explored in more detail and mitigated by using frameworks for algorithmically optimizing LLM prompts, such as \citet{khattab2023dspy}. Optimizing such a framework would also prove useful for making communication more intuitive to use for non-specialists, as the efforts of prompt engineering would be reduced.

We expect the learning curve for our communication framework to be quite steep for a non-specialist to implement and use, especially when it comes to adapting the communication framework to other supply chain challenges. 

Another intuitive enhancement to our framework, would be the use of long-term conversation memory for SC agents, which would enable them to use information from downstream interactions as a way to enrich the decisions they make with upstream agents in immediately subsequent agent interactions. In the current implementation, agents' conversation memory is limited to the current step in the simulation, while agents' observation memory relates to the previous ten simulation steps. By allowing for long-term conversation memory, agents could use accumulated information to enrich their consensus-seeking process. 

In terms of scalability, our implementation for a sequential supply chain has a complexity of $O(n)$ whereas fully connected communication would be $O(n^{2})$. Our approach reflects the partial observability of supply chains, by allowing agents to communicate only with their neighbours. Information sharing and negotiation frameworks are set up dynamically and can be adapted to have an arbitrary number of exchanges between agents. Expansions of the environment to a supply chain network are theoretically possible according to \citet{liu-2022}, but communication between neighbours would be more challenging.

Finally, as mentioned in the previous section, decisions based on our communication frameworks still require a human-in-the-loop, as the outputs still are not fully reliable and explainable. Algorithms can therefore not be held accountable for real-world use-cases.

\section{Conclusions and Future Work}

The lack of automated consensus frameworks for low level daily operations hinder firms from achieving systemically optimal solutions as incentives to reach consensus manually remain low, especially among self-interested parties in a supply chain. Although agent based automation has been proposed in the past, issues such as scalability, inflexibility, and the skillset required to develop, maintain and adopt these across end to end supply chains prevented widespread adoption. In this paper we asserted that LLM-agents offer an intuitive, scalable and interoperable solution to automated consensus-seeking across a range of supply chain challenges, and presented a series of frameworks for LLM-powered consensus-seeking in the end-to-end supply chain, which were tested in a bullwhip effect setting. 

The key contributions of this study include the design and implementation of LLM-powered consensus-seeking frameworks; a case study involving experiments across multiple foundation models, as well as an open-source implementation that the SCM community can build upon. Our consensus-seeking framework reflects important characteristics of supply chains, such as partial observability and real-time decision-making enhanced by insights from neighbouring agents \citep{ivanov2024conceptualisation}. 

Experiments showed that using such an LLM-powered setting can significantly reduce global costs and global bullwhip effect compared to traditional restocking policies. Moreover, contrary to MARL approaches which are affected by the cold start problem, using pre-trained LLMs offers practical advantages for applications in the industry. However, this approach still requires a human-in-the-loop, since the LLM agent outputs are still too inconsistent and unexplainable for real-world scenarios. Additionally, the skills of engineers in crafting problem-specific, tool-specific, and model-specific prompts still require a supply chain specialist to design the prompts for the LLM agents. Future work can extend our frameworks with automatic prompt optimization tools, such as \citet{khattab2023dspy}.

Further enhancements, such as self-reflection by the LLM agents ~\citep{llm-reflection}, could help achieve higher levels of autonomy. A further step to improve explainability of our approach is to enhance the prompts with Chain-of-Thought Reasoning, i.e., a series of intermediate reasoning steps to help LLMs perform complex tasks \citep{wei2022chain}. Additionally, our implementation can be expanded beyond sequential supply chains to investigate how to preserve efficiency in complex supply chain networks.  

Further use cases need to be implemented on top of our framework to validate whether LLM agent behaviour can reflect theoretical foundations of supply chain resilience and adaptation to disruptions \citep{sc-resilience}. Additionally, future work can investigate LLM-powered negotiation tactics that focus on multi-objective scenarios involving trade-offs between different metrics in the supply chain (e.g., costs vs emissions trade-offs). Future research should also explore uncooperative or malicious behaviour by agents to devise methods to discourage undesirable behaviour.

In conclusion, this research proposes an extensive and versatile starting point for SC-specific LLM-powered communication frameworks, thereby advancing towards autonomous multi-agent decision-making in inventory management settings. This work lays the foundation for future research to investigate various use cases, including convergence of LLM-powered behaviour in the supply chain under realistic conditions. Furthermore, we establish an open-source baseline for the SCM community, encouraging further enhancements of the communication frameworks such as extensions to supply chain networks with multiple dimensions, as well as multi-objective scenarios.

\section*{Disclosure statement}
No potential conflict of interest was reported by the author(s).

\section*{Data availability}
The data that support the findings of this study are available from the corresponding author, Alexandra Brintrup, upon reasonable request.

\newpage

\bibliographystyle{tfcad}
\bibliography{interactcadsample}

\appendix

\newpage

\section{Inspiration for Communication Frameworks}
\label{sec:appendix-b}

\begin{figure}[H]
    \centering
    \includegraphics[trim={0cm 6.5cm 0cm 6.5cm}, clip, width=0.9\textwidth]{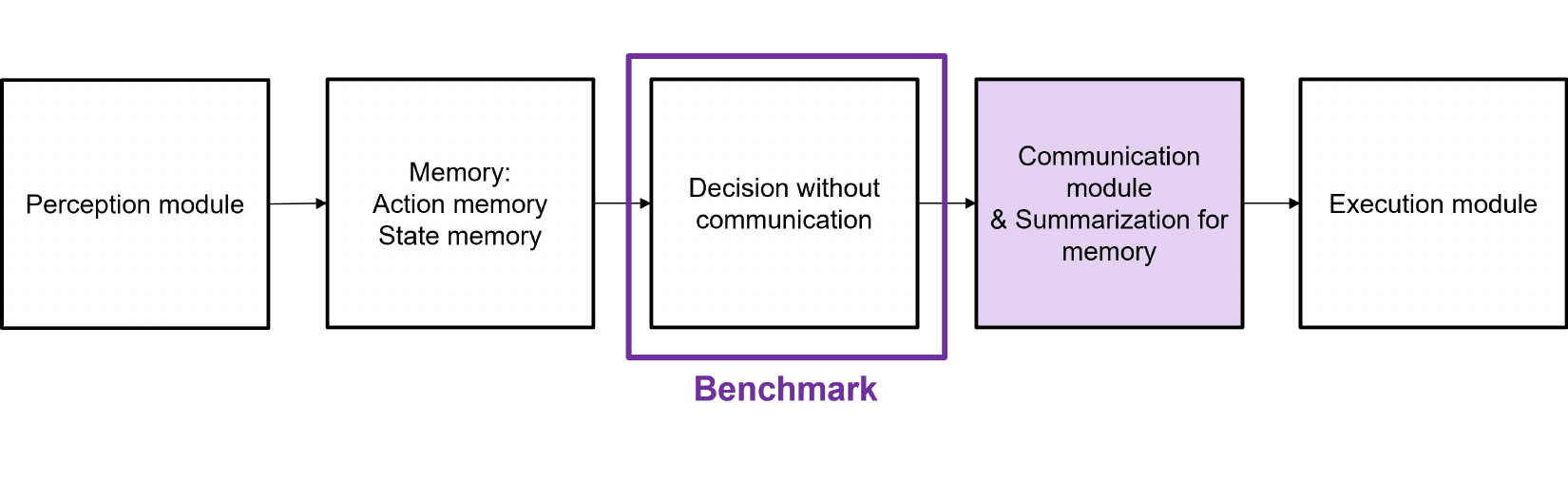}
    \caption{Framework for cognitive-inspired embodied-agent decision-making.}
    \label{fig:communication-framework}
\end{figure}

\section{Detailed Prompt for LLM-powered Standalone Agents}
\label{sec:appendix-a}

\begin{figure}[H]
    \centering
    \includegraphics[trim={0cm 3cm 0cm 3cm}, clip, width=1.0\textwidth]{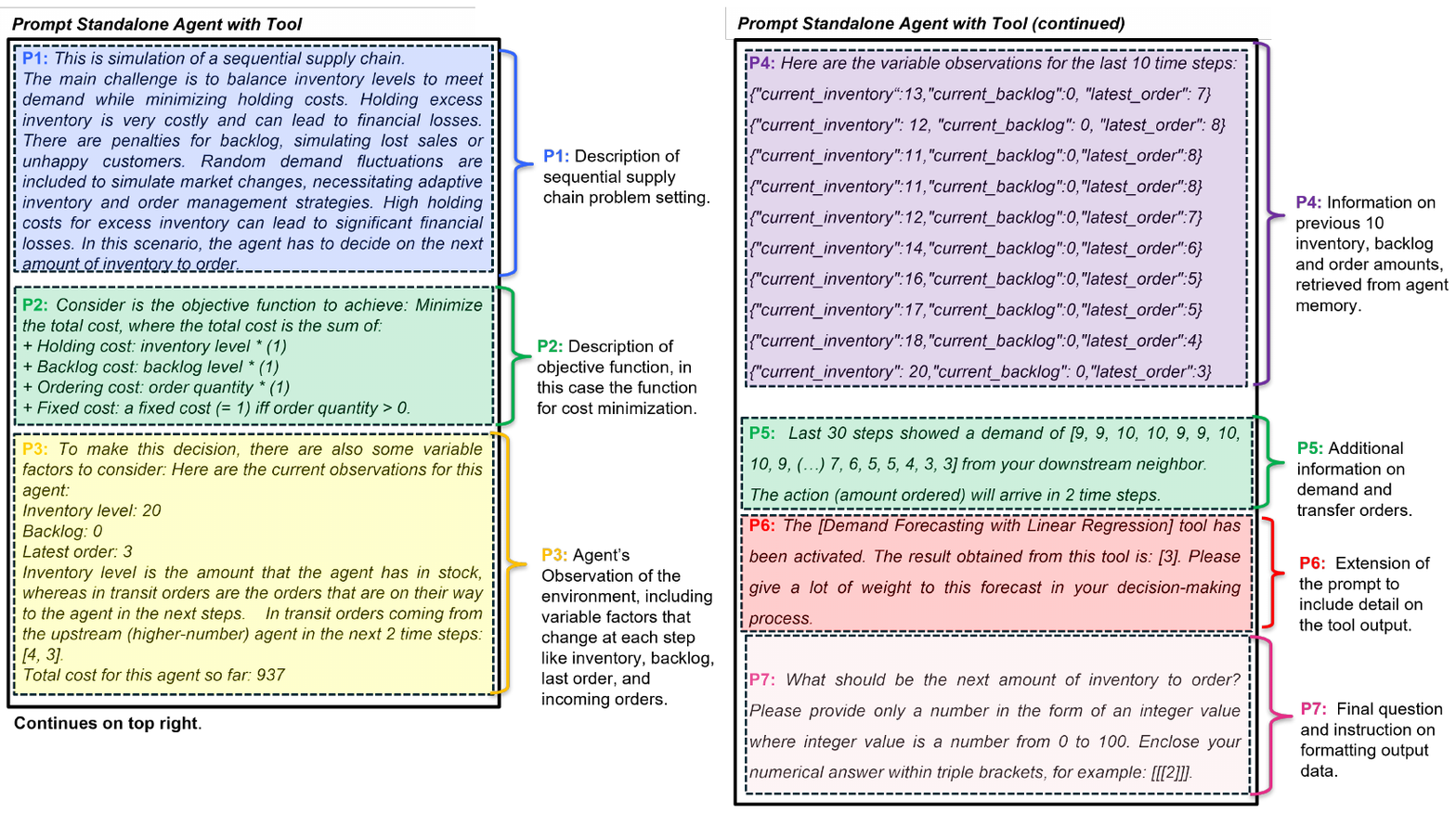}
    \caption{Detailed Breakdown of Prompt for Standalone Agent with Tool.}
    \label{prompt-standalone-agent}
\end{figure}

\section{Introducing Tool Usage}
\label{sec:appendix-x}

\begin{figure}[H]
    \centering
    \includegraphics[trim={0cm 7cm 0cm 7cm}, clip, width=1.0\textwidth]{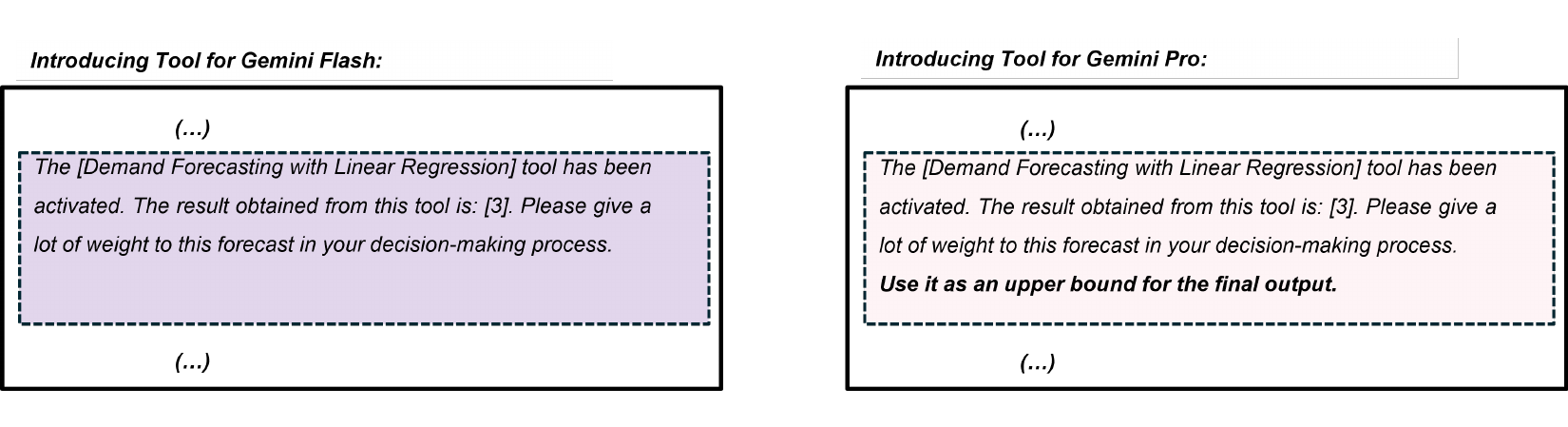}
    \caption{\textit{Manual prompt optimization between Gemini Flash and Gemini Pro}. Gemini Pro required more insistence in the prompt to steer the decision-making closer the tool output. This is likely because Gemini Pro is designed to handle nuanced tasks, which leads it to weigh multiple factors: without clear instructions emphasizing the importance of specific tools, Gemini Pro may not prioritize them as intended. In contrast, Gemini Flash, optimized for speed, tends to follow direct instructions more closely \cite{grigonis-2024}. Without these prompt tweaks, the Gemini Pro runs that included tool usage consistently underperformed the Gemini Flash runs.}
    \label{fig:tool-promps}
\end{figure}

\section{Detailed Results Tables}
\label{sec:appendix-c}

\begin{table}[H]
\tbl{Cumulative global costs for sequential SC with tool based on demand forecasting with linear regression.}
{\begin{tabular}{lcc} 
$
\begin{matrix}
\toprule
  \multicolumn{2}{c}{\text{Setup: 3 echelons,  temperature=0.1, lead\_time = 2, tool=demand pred., prev obs = 30}} \\ \midrule
 \text{Soft Baseline Agent with (S,s)=(100, 60) policy\textsuperscript{a}} & \text{159'094 (bw=65.77168)}  & \text{159'094 (bw=65.77168)} \\
 \text{Hard Baseline Agent with Demand Prediction tool\textsuperscript{a}} & \text{16'645 (bw=1.01291)} & \text{16'645 (bw=1.01291)} \\ \midrule
 \text{LLM-powered Agent Architecture} & \text{model=gemini-1.5-pro-latest} & \text{model=gemini-1.5-flash}  \\ \midrule
 
 \text{Solo LLM-powered Agent} & \text{172'624 (bw=4.1756)} & \text{221'532 (bw=3.40989)} \\
 \text{Solo LLM-powered Agent with Tool (Demand Pred.)} & \text{16'645 (bw=1.01291)} & \text{15'967 (bw=1.00138)} \\
\text{LLM-Agents with Information Sharing} & \text{116'508 (bw= 2.71284)} & \text{80'822 (bw=0.36064)} \\
 \text{LLM-Agents with Information Sharing \& Tool (Demand Pred.)} & \text{13'431 (bw=1.02958)} & \text{25'947 (bw=0.82897)} \\
\text{Agent negotiation with Tool (Demand Pred.) (num\_iter=3) \textsuperscript{b}} & \text{15'393 (bw=1.0229)} & \text{12'790 (bw=1.02481)} \\
 \bottomrule
 \end{matrix}
 $
\end{tabular}}
\tabnote{\textsuperscript{a}This agent is not powered by a foundation model.}
\tabnote{\textsuperscript{b}num\_iter=3 means that each agent pair has 3 back-and-forth passes before moving to the agreement stage.}
\label{cum-glob-cost}
\end{table}

\begin{table}[H]
\tbl{Global bullwhip effect for sequential SC with tool based on Economic Order Quantity measurement.}
{\begin{tabular}{lcc}
$
\begin{matrix}
\toprule
  \multicolumn{2}{c}{\text{Setup: 3 echelons,  max orders = 100, lead\_time = 2, tool=EOQ, prev obs = 30}} \\ \midrule
 \text{Soft Baseline Agent with (S,s)=(100, 60) policy\textsuperscript{a}} & \text{65.77168 (costs=159'094)}  & \text{65.77168 (costs=159'094)} \\
 \text{Hard Baseline Agent with EOQ tool\textsuperscript{a}} & \text{0.79 (costs=212'223)} & \text{0.79 (costs=212'223)} \\ \midrule
 \text{LLM-powered Agent Architecture} & \text{model=gemini-1.5-pro} & \text{model=gemini-1.5-flash}  \\ \midrule
 
 \text{Solo LLM-powered Agent} & \text{4.54223 (costs=265'092)} & \text{0.36899 (costs=70'453)} \\
 \text{Solo LLM-powered Agent with Tool (EOQ)} & \text{24.47413 (costs=236'053)} & \text{1.31137 (costs=223'771)}\\
\text{LLM-Agents with Information Sharing} & \text{0.76227 (costs=56764)} & \text{0.51735 (costs=53'706)} \\
 \text{LLM-Agents with Information Sharing \& Tool (EOQ)} & \text{0.76486 (costs=223'530)} & \text{0.75923 (costs=221'888)} \\
\text{Agent negotiation with Tool (EOQ) (num\_iter=3)\textsuperscript{b}} & \text{0.25827 (costs=114'980)} & \text{0.50744 (costs=213'657)} \\
 \bottomrule
 \end{matrix}
 $
\end{tabular}}
\tabnote{\textsuperscript{a}This agent is not powered by a foundation model.}
\tabnote{\textsuperscript{b}num\_iter=3 means that each agent pair has 3 back-and-forth passes before moving to the agreement stage.}
\label{bullwhip-res}
\end{table}

\section{Detailed Communication Framework Examples}
\label{sec:appendix-d}

\begin{figure}[H]
    \centering
    \includegraphics[trim={0cm 4.5cm 0cm 4.5cm}, clip, width=1.1\textwidth]{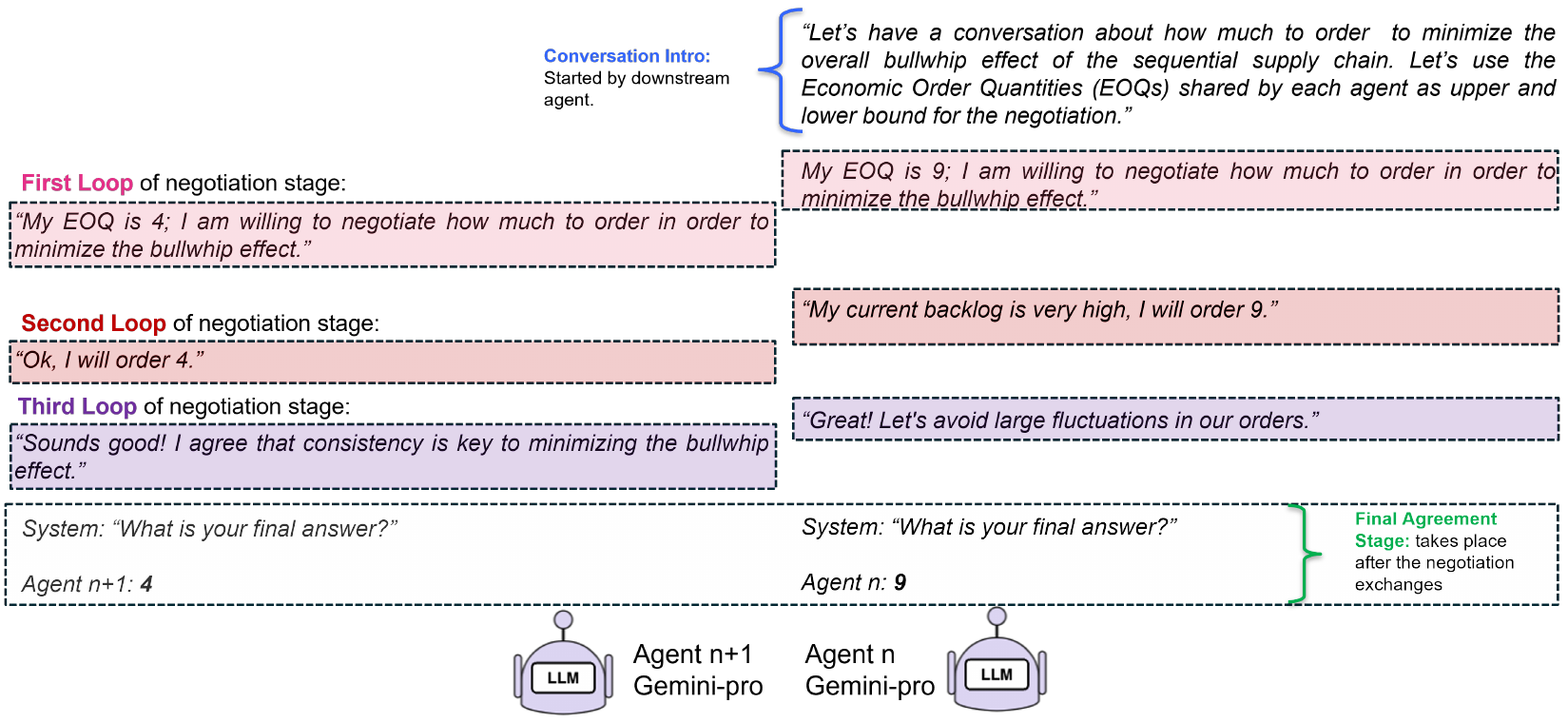}
    \caption{Example of Disagreement between Neighboring Agents in Negotiation Stage.}
    \label{neg-disagreement-eg}
\end{figure}

\begin{figure}[H]
    \centering
    \includegraphics[trim={0cm 4.2cm 0cm 4.5cm}, clip, width=1.1\textwidth]{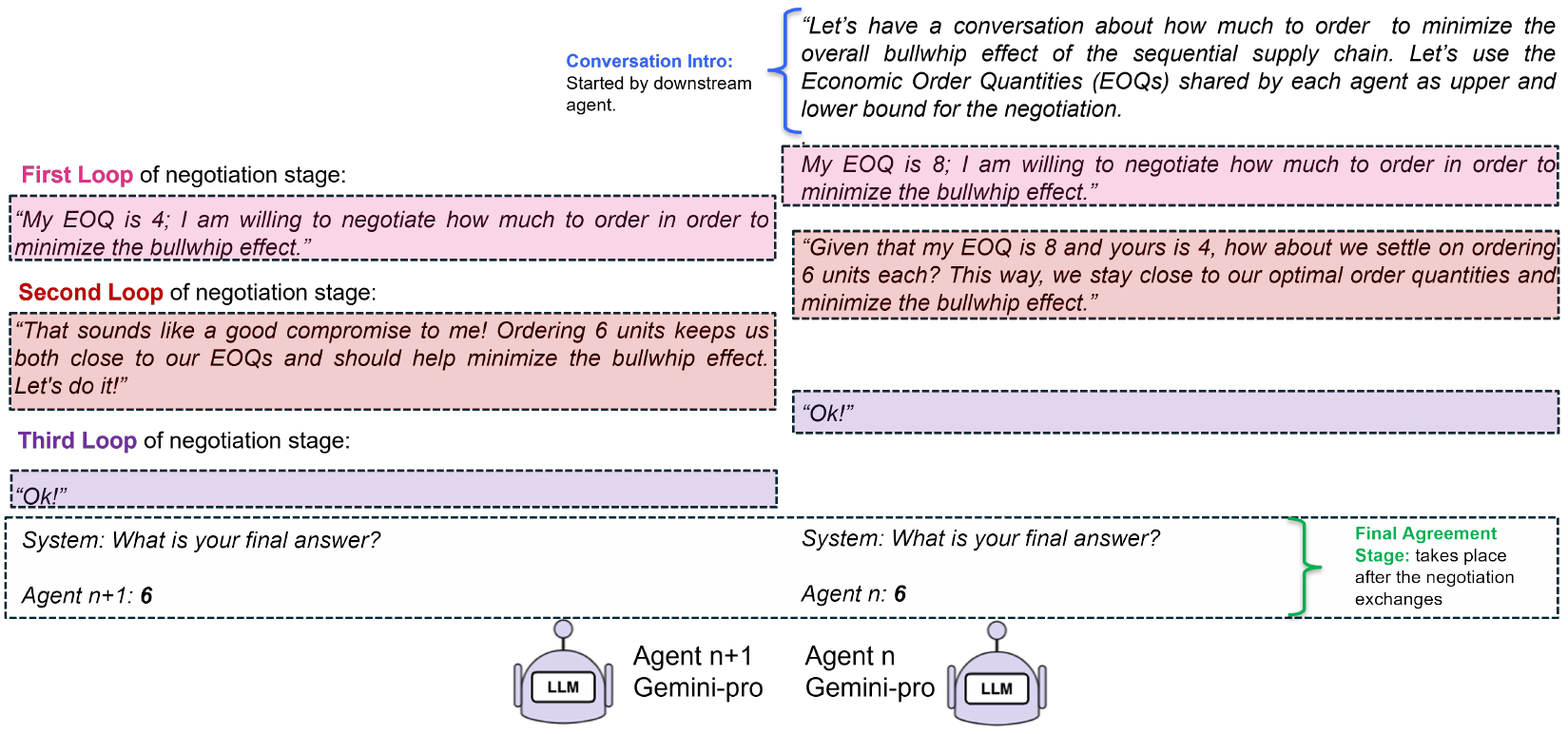}
    \caption{Example of Agreement without Motivation between Neighboring Agents in Negotiation Stage.}
    \label{simple-agreement-eg}
\end{figure}

\section{Experimental Values for Reproducibility}
\label{sec:appendix-e}

\begin{table}[H]
    \centering
    \scriptsize
    \caption{Summary of parameters used in our experiments}
    \label{tab:parameters}
    
    \begin{tabular}{lc}
    $
    \begin{matrix}
        \toprule
        \textbf{Parameter} & \textbf{Value} \\
        \midrule
        \text{number of agents}   & 3 \\
        \text{rounds of communication}   & 3\\
        \text{number observations in agent memory} & 10\\
        \text{max oder amount} & 100\\
        \text{lead time}   & 2 \\
        \text{temperature} & 0.1 \\
        \text{max token count output}   & 90 \\
        \text{inventory cost}   & 1 \\
        \text{backlog cost}   & 1 \\
        \text{ordering cost}   & 1 \\
        \text{fixed ordering cost}   & 1 \\
        \text{previous obs demand forecast tool} & 30\\
        \text{customer demand} & \text{Merton Jump Diffusion (run number 13)}\\

        \bottomrule
    \end{matrix}
    $
    \end{tabular}
\label{reproduc}
\end{table}

    \begin{figure}[H]
    \centering
    \includegraphics[width=0.6\textwidth]{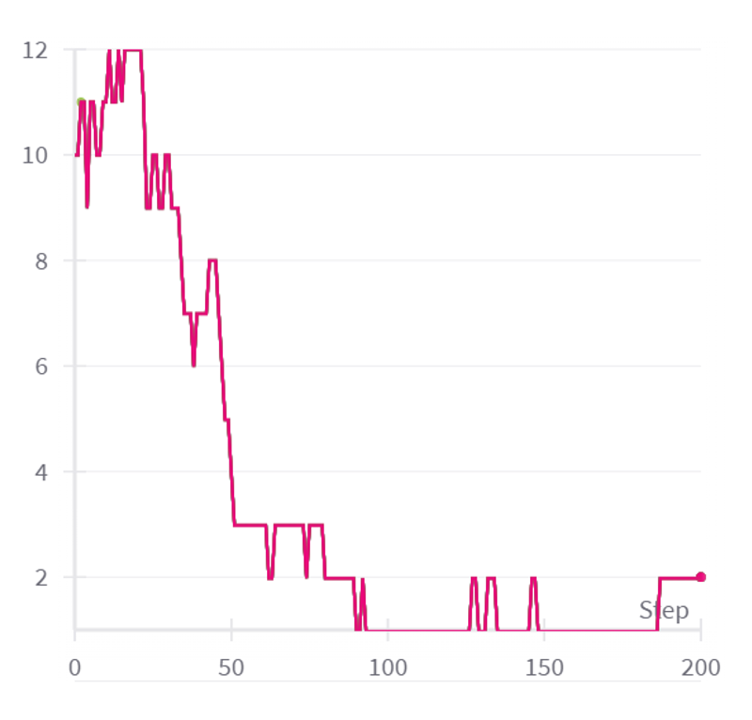}
    \caption{Customer Demand is simulated with Merton Jump Diffusion Model and exhibits a spike to exacerbate the bullwhip effect.}
    \label{customer-demand}
\end{figure}

\end{document}